\documentclass[a4paper,12pt]{article}

\usepackage{float}
\usepackage{amsmath} 
\usepackage{amssymb}
\usepackage{amsfonts}
\usepackage{epstopdf}
\usepackage{color,fancybox,graphicx}
\usepackage{url}
\usepackage{psfrag}
\usepackage{color}
\usepackage{pifont}

\usepackage{pstricks, pst-tree}

\numberwithin{equation}{section}

\newtheorem{rem}{Remark}[section]

\newtheorem{lem}{Lemma}[section]
\newtheorem{cor}{Corollary}[section]
\newtheorem{pro}{Proposition}[section]
\newtheorem{theo}{Theorem}[section]
\newtheorem{fact}{Fact}[section]

\newcommand{\bX}{\mathbf{X}}
\newcommand{\bx}{\mathbf{x}}
\newcommand{\bz}{\mathbf{z}}
\definecolor{gris25}{gray}{0.90}

\usepackage{natbib}
\bibpunct{(}{)}{;}{a}{,}{,}

\begin{document}

\begin{center}

{\Large 
\textbf{\textsf{Cellular Tree Classifiers}}}
\medskip

\medskip

\end{center}
{\bf G\'erard Biau\\
{\it Universit\'e Pierre et Marie Curie\footnote{Research partially supported by the French National Research Agency (grant ANR-09-BLAN-0051-02 ``CLARA'') and by the Institut universitaire de France.} \& Ecole Normale Sup{\'e}rieure\footnote{Research carried out within the INRIA project ``CLASSIC'' hosted by Ecole Normale Sup{\'e}rieure and CNRS.}, France}}\\
\textsf{gerard.biau@upmc.fr}
\bigskip

{\bf Luc Devroye\\
{\it McGill University, Canada}\footnote{Research sponsored by NSERC Grant A3456 and FQRNT Grant 90-ER-0291.}}\\
\textsf{lucdevroye@gmail.com}

\medskip

\begin{abstract}
\noindent {\rm The cellular tree classifier model addresses a fundamental problem
in the design of classifiers for a parallel or distributed computing
world: Given a data set, is it sufficient to apply a majority rule
for classification, or shall one split the data into
two or more parts and send each part to a potentially different
computer (or cell) for further processing?
At first sight, it seems impossible to define with this paradigm a consistent classifier as no cell knows
the ``original data size'', $n$. 
However, we show that this is not so by exhibiting
two different consistent classifiers.
The consistency is universal but is only shown for
distributions with nonatomic marginals.

\medskip

\noindent \emph{Index Terms} --- 
Classification,
pattern recognition,
tree classifiers,
cellular computation,
Bayes risk consistency,
asymptotic analysis,
nonparametric estimation.
\medskip

\noindent \emph{2010 Mathematics Subject Classification}: 62G05, 62G20.}

\end{abstract}

\section{Introduction}
\subsection{The problem}
We explore in this paper a new way of dealing with the supervised classification problem. In the model we have in mind, a basic computational unit in classification, a cell, 
takes as input training data, and makes a decision
whether a majority rule should be applied to all data,
or whether the data should be split, and each part of the
partition should be given to another cell.
All cells must be the same---their function
is not altered by external inputs. In other words,
the decision to split depends only upon the data
presented to the cell. Classifiers designed according to this autonomous principle will be
called cellular tree classifiers, or simply cellular classifiers. 
This manner of tackling the classification
problem is novel, but has a wide reach in a world in
which parallel and distributed computation 
are important. In the short term, parallelism will take hold in massive data sets and complex systems and, as such, is one of the exciting questions that will be asked to the statistics and machine learning fields.
\medskip

The purpose of the present document is to formalize the setting and to provide a foundational discussion of various properties, good and bad, of tree classifiers that are formulated following these principles. Our constructions lead to classifiers that always converge.
They are the first consistent
cellular classifiers that we are aware of. 
This article is also motivated by the challenges involved in ``big data'' issues \citep[see, e.g.,][]{Jordan}, in which recursive approaches such as divide-and-conquer algorithms \citep[e.g.,][]{CLRS} play a central role. Such procedures are naturally adapted for execution in multi-processor machines, especially shared-memory systems where the communication of data between processors does not need to be planned in advance.
\medskip

In the design of classifiers, we have an unknown distribution of a random prototype pair $(\bX,Y)$, where $\bX$ takes values in $\mathbb R^d$ and $Y$ takes only finitely many values, say 0 or 1 for simplicity. Classical pattern recognition deals with predicting the unknown nature $Y$ of the observation $\bX$ via a measurable classifier $g:{\mathbb R}^d\to\{0,  1\}$. Since it is not assumed that $\bX$ fully determines the label, it is certainly possible to misspecify its associated class. Thus, we err if $g (\bX)$
differs from $Y$, and the probability of error for a particular decision rule $g$ is  
$L (g)=\mathbb P\{ g (\bX)\neq Y\}$. 
The Bayes classifier
\begin{eqnarray*} 
g^{\star} (\bx) = \left\{
\begin{array}{ll}
1 & \mbox{ if  $\mathbb P\{Y=1| \bX=\bx\} > \mathbb P\{Y=0| \bX=\bx\}$}\\
0 & \mbox{ otherwise}
\end{array}
\right.
\end{eqnarray*} 
has the smallest
probability of error, that is
\begin{equation*}
L^{\star} = L (g^{\star}) = \inf_{g: \mathbb R^d \to \{0,  1\} } \mathbb P\{g(\bX)\neq Y\}
\end{equation*}
\citep[see, for instance, Theorem 2.1 in][]{DGL}. However, most of the time, the distribution of $(\bX,Y)$ is unknown, so that $g^{\star}$ is unknown too.  Fortunately, it is often possible to collect a sample (the data) $\mathcal D_n=((\bX_1,Y_1), \hdots, (\bX_n,Y_n))$ of independent and identically distributed (i.i.d.) co\-pies of $(\bX,Y)$. We assume that $\mathcal D_n$ and $(\bX,Y)$ are independent.  In this context, a classifier $g_n(\bx;\mathcal D_n)$ is a measurable function of $\bx$ and $\mathcal D_n$, and it attempts to estimate $Y$ from $\bX$ and $\mathcal D_n$. For simplicity, we suppress $\mathcal D_n$ in the notation and write $g_n(\bx)$ instead of $g_n(\bx;\mathcal D_n)$. 
\medskip

The probability of error of a given classifier $g_n$ is the random variable
$$L(g_n) = \mathbb P\{g_n(\bX)\neq Y | \mathcal D_n\},$$
and the rule is consistent if
$$\lim_{n \to \infty} \mathbb E L(g_n)=L^{\star}.$$
It is universally consistent if it is consistent for all possible distributions of $(\bX,Y)$. Many popular
classifiers are universally consistent. These include several brands of histogram rules, $k$-nearest neighbor
rules, kernel rules, neural networks, and tree classifiers. There are too many references to be cited here, but the monographs by \citet{DGL} and \citet{Laciregressionbook} will provide the reader with a comprehensive introduction to the domain and a literature review. Among these rules, tree methods loom large for
several reasons. 
All procedures that partition space, such as histogram rules, can be viewed as special cases
of partitions generated by trees. Simple neural networks that use voting methods can also be regarded as
trees, and similarly, kernel methods with kernels that are indicator functions of sets are but special cases
of tree methods. Tree classifiers are conceptually simple, and explain the data very well. However, their design can
be cumbersome, as optimizations performed over all possible tree classifiers that follow certain restrictions
could face a huge combinatorial and computational hurdle. The 
cellular paradigm addresses these concerns.
\medskip

Partitions of $\mathbb R^d$ based upon trees
have been studied in the computational geometry
literature \citep[][]{Bentley75, Overmars82,Edelsbrunner83,Mehlhorn84}
and the computer graphics literature \citep[][]{Samet84a,Samet90a}. 
Most popular among these are the $k$-$d$ trees and quadtrees.
Our version of space partitioning corresponds to
Bentley's $k$-$d$ trees \citeyearpar{Bentley75}.
The basic notions of trees as related
to pattern recognition can be found in Chapter 20
of \citet{DGL}.
However, trees have been suggested
as tools for classification more than twenty years before that.
We mention in particular the early work of Fu \citep[][]{YoFu76,AnFu79,MuFu80,LiFu83,QiFu83}.
Other references from the 1970s include \citet{Meisel73, Bartolucci76,PaMe77,SeCh77,Swain77,Gordon78,Fri79}.
Most influential in the classification tree
literature was the CART proposal
by \citet{Breimanbook}.
While CART proposes partitions by
hyperrectangles, 
linear hyperplanes in general position
have also gained in popularity---the early work on that topic is
by \citet{LoVa88}, and \citet{PaSk90}.
Additional references on tree classification include
\citet{Gustafson80,ArChBe82,HaVaMeGe82,Kurzynski83,WaSu84,SuWa87,Shl90,Cho91,GeDe91,Gelfand91,Sim91,GuGe92}.
\subsection{The cellular computation spirit}
In general, classification trees partition $\mathbb R^d$ into regions, often hyperrectangles
parallel to the axes (an example is depicted in Figure \ref{figure1}). In $t$-ary trees, each node has exactly $t$ or $0$ children. If a node $u$ represents the set $A$ and its children 
$u_1, \hdots, u_t$ represent $A_1, \hdots, A_t$, then it is required that $A=A_1\cup\cdots \cup A_t$ and $A_i \cap A_j=\emptyset$ for $i\neq j$. The root
of the tree represents $\mathbb R^d$, and the terminal nodes (or leaves), taken together, form a partition of $\mathbb R^d$. If a leaf represents region $A$, then the tree classifier takes the simple form
\begin{eqnarray*} 
g_n (\bx) = \left\{
\begin{array}{ll}
1 & \mbox{ if  $\sum_{i=1}^n \mathbf 1_{[\bX_i \in A, Y_i =1]} > \sum_{i=1}^n \mathbf 1_{[\bX_i \in A, Y_i =0]}$, \quad $\bx\in A$}\\
0 & \mbox{ otherwise.}
\end{array}
\right.
\end{eqnarray*} 
That is, in every leaf region, a majority vote is taken over all $(\bX_i,Y_i)$'s with $\bX_i$'s in the same region. Ties are broken, by convention, in favor of class 0.
\begin{figure}[!t]
\centering
\begin{minipage}[c]{.46\linewidth}
\psset{treemode=D,levelsep=35pt}
\pstree{\Tcircle{}}
{

	\pstree{ \Tcircle{} }  
		{
		\pstree{ \Tr{\psframebox{\textsf{leaf}}} }{}
		\pstree{ \Tr{\psframebox{\textsf{leaf}}} }{}
		}
	\pstree{ \Tcircle{} }
	{
			\pstree{ \Tcircle{} }
			{
			
				\pstree{ \Tr{\psframebox{\textsf{leaf}}} }{}
				\pstree{ \Tcircle{} }{
				
				\pstree{ \Tr{\psframebox{\textsf{leaf}}} }{}
				\pstree{ \Tr{\psframebox{\textsf{leaf}}} }{}
				
				}
			}
			\pstree{ \Tr{\psframebox{\textsf{leaf}}} }{}
	}

	}
\end{minipage} 
\begin{minipage}[c]{.46\linewidth}
\begin{pspicture}(0,0)(6,6)
\psline(2,0)(2,6)
\psline(0,4)(2,4)
\psline(2,2.5)(6,2.5)
\psline(3,0)(3,2.5)
\psline(3,1)(6,1)
\pscircle*(3.45,1.23){0.1}
\pscircle(3.5,3.77){0.1}
\pscircle*(1.2,4.78){0.1}
\pscircle(0.5,1.12){0.1}
\pscircle*(1.6,4.88){0.1}
\pscircle(2.8,5.97){0.1}
\pscircle*(1.2,0.22){0.1}
\pscircle(0.85,0.78){0.1}
\pscircle*(1.8,1.33){0.1}
\pscircle(2.3,3.4){0.1}
\pscircle*(5.56,2.89){0.1}
\pscircle(0.43,3.237){0.1}
\pscircle*(5.78,4.89){0.1}
\pscircle(1.23,5.78){0.1}
\pscircle*(0.12,5.23){0.1}
\pscircle(3.89,5.27){0.1}
\pscircle*(5.12,3.2){0.1}
\pscircle(0.25,4.89){0.1}
\pscircle*(4.23,5.11){0.1}
\pscircle(3.89,2.25){0.1}
\pscircle*(4.12,1.78){0.1}
\pscircle(5.98,1.22){0.1}
\pscircle*(1.23,3.81){0.1}
\pscircle(3.37,2.78){0.1}
\pscircle*(4.78,1.87){0.1}
\pscircle(5.5,1.99){0.1}
\pscircle*(2.3,2.12){0.1}
\pscircle(2.95,3.7){0.1}
\pscircle*(3.68,3.229){0.1}
\pscircle(3.35,3.22){0.1}
\pscircle*(4.2,3.78){0.1}
\pscircle(3.78,4.52){0.1}
\pscircle(1.48,1.2){0.1}
\pscircle(2.78,2.62){0.1}
\pscircle(0.99,1.92){0.1}
\pscircle(0.24,3.85){0.1}
\pscircle(1.78,3.62){0.1}
\pscircle(1.81,4.19){0.1}
\pscircle(1.118,2.90){0.1}
\pscircle(2.378,1.362){0.1}
\pscircle(2.84,0.1275){0.1}
\pscircle(1.78,3.62){0.1}
\pscircle(4.81,1.19){0.1}
\pscircle(5.118,0.81){0.1}
\pscircle(5.718,1.52){0.1}
\pscircle*(3.78,1.62){0.1}
\pscircle*(4.121,0.2){0.1}
\pscircle*(5.118,0.81){0.1}
\pscircle*(5.178,1.229){0.1}
\pscircle*(2.621,4.65){0.1}
\pscircle(5.221,0.25){0.1}
\end{pspicture}
\end{minipage}
\caption{\label{figure1}\textsf{A binary tree (left) and the corresponding partition (right).}}
\end{figure}
\medskip

The tree structure is usually data-dependent, as well, and indeed, it is in the construction itself where different trees differ. Thus, there are virtually infinitely many possible strategies to build classification trees. Nevertheless, despite this great diversity, all tree species end up with two fundamental questions at each node:
\medskip

\begin{center}
\fcolorbox{black}{gris25}{
\begin{minipage}{0.6\textwidth}
\begin{enumerate}
\item[\ding{172}] \textsf{Should the node be split?}
\item[\ding{173}] \textsf{In the affirmative, what are its children?}
\end{enumerate}
\end{minipage}
}
\end{center}
\medskip

These two questions are typically answered using {\bf global} information regarding the tree, such as, for example, a function of the data $\mathcal D_n$, the level of the node within the tree, the size of the data set and, more generally, any parameter connected with the structure of the tree. This parameter could be, for example, the total number $k$ of cells in a $k$-partition tree or the penalty term in the pruning of the CART algorithm (\citealp[][]{Breimanbook}; see also \citealp{Servane}). 
\medskip

Cellular trees proceed from a different philosophy.  In short, a cellular tree should, at each node, be able to answer questions \ding{172} and \ding{173} using {\bf local} information only, without any help from the other nodes. In other words, each cell can perform as many operations it wishes, provided it uses {\bf only} the data that are transmitted to it, regardless of the general structure of the tree. Just imagine that the calculations to be carried out at the nodes are sent to different computers, eventually asynchronously, and that the system architecture is so complex that computers do not communicate. 
Such a situation may arise, for example, in the context of massive data sets, that is, when both $n$ and $d$ are astronomical, and no single human and no single computer can handle this alone. Thus, once a computer receives its data, it has to make its own decisions \ding{172} and \ding{173} based on this data subset only, independently of the others and without knowing anything of the overall edifice. Once a data set is split, it can be given to another computer for further splitting, since the remaining data points have no influence. This greedy mechanism is schematized in Figure \ref{figure2}.
\medskip

But there is a more compelling reason for making local decisions. A neurologist 
seeing twenty patients must make decisions without knowing anything about the
other patients in the hospital that were sent to other specialists. Neither
does he need to know how many other patients there are. The neurologist's
decision, in other words, should only be based on the data---the patients---in his
care. 
\pstree[edge=none]{\Tp{}}

\begin{figure}[!t]
\centering
$\begin{psmatrix}[colsep=0.2]
 & \begin{pspicture}(-0.2,-0.2)(1,1)
\pscircle(0.56,0.90){0.1}
\pscircle(0.5,0.062){0.1}
\pscircle*(0.458,0.362){0.1}
\pscircle(0.93,0.48){0.1}
\pscircle*(0.12,0.52){0.1}
\pscircle*(0.98,0.85){0.1}
\pscircle(0.078,0.12){0.1}
\pscircle*(0.9,0.1){0.1}
\pscircle(0,0.79){0.1}
\end{pspicture} & &  \psframebox{\mbox{\textsf{Majority vote}}}\\
& \psframebox{\mbox{\textsf{Cellular decision}}} & &\psframebox{\mbox{\textsf{No split}}}\\
 &\psframebox{\mbox{\textsf{Split}}} & &\\
\begin{pspicture}(0.2,0.2)(1,1)
\pscircle(0.56,0.90){0.1}
\pscircle(0.5,0.062){0.1}
\pscircle*(0.458,0.362){0.1}
\pscircle(0.93,0.48){0.1}
\pscircle*(0.12,0.52){0.1}
\end{pspicture} & &
\begin{pspicture}(0.2,0.2)(1,1) 
\pscircle*(0.98,0.85){0.1}
\pscircle(0.178,0.12){0.1}
\pscircle*(0.9,0.1){0.1}
\pscircle(0.43,0.79){0.1}
\end{pspicture} & 
\end{psmatrix}$
\psset{arrowscale=2,arrows=->}
\ncline{1,2}{2,2}
\ncline{2,2}{3,2}
\ncline{2,2}{2,4}
\ncline{3,2}{4,1}
\ncline{3,2}{4,3}
\ncline{2,4}{1,4}
\caption{\label{figure2}\textsf{Schematization of the cell, the computational unit.}}
\end{figure}
\vspace{-1.5cm}

Decision tree learning is a method commonly used in data mining
\citep[see, e.g.,][]{RoMa}. 
Its goal is to create a model that partitions the space recursively, as 
in a tree, in which leaf nodes (terminal nodes) correspond
to final decisions.  
This process of top-down induction of decision trees---a phrase
introduced by Quinlan in 1968---is called greedy in the
data mining and computer science literature.
It is by far the most common strategy for learning decision trees from data.
The literature on this topic is largely concerned with 
the manner in which splits are made, and
with the stopping rule.  
\medskip

For example, in CART \citep{Breimanbook},
splits are made perpendicular to the axes based on the notion of Gini impurity.
Splits are performed until all data are isolated.
In a second phase, nodes are recombined from the bottom-up 
in a process called pruning.  It is this second process that makes
the CART trees non-cellular, as global information is
shared to manage the recombination process.
Quinlan's C4.5 \citeyearpar{Qu93} also prunes.
Others split until all nodes or cells are homogeneous (i.e., have
the same class)---the prime example is Quinlan's ID3 \citeyearpar{Qu86}. 
This strategy, while compliant with the 
cellular framework, leads to non-consistent rules, 
as we point out in the present paper.
In fact, the choice of a good stopping rule for
decision trees is very hard---we were not able to find any
in the literature that guarantee convergence
to the Bayes error.
\medskip

We note here that decision networks have received renewed attention
in wireless sensor networks (\citealp[see, e.g.,][]{Arora}, or \citealp[][]{Cheng}). 
Physical and energy considerations
impose a natural restriction on the classifiers---decisions must be taken
locally.  This corresponds, in spirit, to the cellular
framework we are proposing. However, most sensor network decision trees use
global criteria such as pruning that are based on a global method of deciding
where to prune.  The consistency question has not been addressed in
these applications.
\section{Cellular tree classifiers}

\subsection{A mathematical model}
The objective of this subsection is to discuss a tentative mathematical model for cellular tree classifiers. 
Without loss of generality, we consider binary tree classifiers based on a class $\mathcal C$ of possible Borel subsets of $\mathbb R^d$ that can be used for splits. A typical example of such a class is the family of all hyperplanes,
or the class of all hyperplanes that are perpendicular to one
of the axes. Higher order polynomial splitting surfaces can
be imagined as well. 
\medskip

The class is parametrized by a vector $\sigma \in \mathbb R^p$. 
There is a splitting function $f(\bx,\sigma)$, $\bx \in \mathbb R^d, \sigma \in \mathbb R^p$, such that
$\mathbb R^d$ is partitioned into
$A = \{ \bx \in \mathbb R^d: f(\bx,\sigma) \ge 0 \}$
and $B = \{ \bx \in \mathbb R^d: f(\bx,\sigma) < 0 \}$.
Formally, a cellular split can be viewed
 as a family of measurable mappings $\sigma$ from $(\mathbb  R^d \times \{0,1\})^n$ to
$\mathbb R^p$
(for all $n \geq 1$). That is, for each possible input size $n$,
we have a map.
In addition, there is a family of measurable mappings $\theta$ from $(\mathbb  R^d \times \{0,1\})^n$ to
$\{0,1\}$ that
indicate decisions: $\theta = 1$ indicates that a split should be applied,
while $\theta = 0$ corresponds to a decision not to split. In that
case, the cell acts as a leaf node in the tree.
Note that $\theta$ and $\sigma$ correspond to the decisions
given in \ding{172} and \ding{173}.
\medskip

A cellular binary classification tree is a machine that partitions the space recursively in the following
manner. With each node we associate a subset of $\mathbb R^d$, starting with $\mathbb R^d$ for the root node. 
Let the data set be $\mathcal D_n$.
If $\theta (\mathcal D_n) =  0$, the root cell is final, and the space is
not split. Otherwise, $\mathbb R^d$ is split into
$$
A = \left \{ \bx \in \mathbb R^d: f\left (\bx, \sigma(\mathcal D_n) \right) \ge 0 \right \} \quad \mbox{and} \quad
B = \left \{ \bx \in \mathbb R^d: f\left (\bx, \sigma(\mathcal D_n) \right) < 0 \right \}.
$$
The data $\mathcal D_n$ are partitioned into two groups---the first group contains all
$(\bX_i, Y_i)$, $i=1, \hdots, n$, for which $\bX_i \in A$, and the second group
all others.  The groups are sent to child cells,
and the process is repeated.
\medskip

A priori, there is no reason why this tree should be
finite. We will impose conditions later on that ensure that with probability 1, the tree is finite for all $n$ and for all possible values of the data.
For example, this could be achieved by hyperplane splits perpendicular to the axes
that are forced to visit (contain) one of the $\bX_i$'s. 
By insisting that the data point selected on the boundary be ``eaten'', i.e.,
not sent down to the child nodes, one reduces the data set by one at each split,
thereby ensuring the finiteness of the decision tree.
We will employ such a (crude) method.
\medskip

When $\bx \in \mathbb R^d$ needs to be classified, we
first determine
the unique leaf set $A(\bx)$ to which $\bx$ belongs, and then take votes among the $\{Y_i : \bX_i \in A(\bx), i=1, \hdots, n\}$.
Classification proceeds by a majority vote,
with the majority deciding the estimate $g_n(\bx)$. In case of a tie, we set $g_n(\bx)=0$. 
\medskip

A cellular binary tree classifier is said to be randomized if each node in the tree has an independent copy
of a uniform $[0,1$] random variable associated with it, and $\theta$ and $\sigma$ are mappings that have one extra
real-valued component in the input. For example, we could flip an unbiased coin at each node to decide
whether $\theta=0$ or $\theta= 1$.
\begin{rem}
\label{Remark2}
It is tempting to say that any classifier $g_n$ 
is a cellular tree classifier with the following mechanism: 
Set $\theta=1$ if we are at the root, and $\theta=0$ elsewhere. 
The root node is split by the classifier into a set 
$$A=\{\bx \in \mathbb R^d : g_n(\bx) = 1\}$$
and its complement, and both child nodes are leaves. 
However, the decision to cut can only be a function of the input
data, and not the node's position in the tree, and thus, this is not allowed. 
\end{rem}
\subsection{Are there consistent cellular tree classifiers?}
At first sight, it appears that there are no universally consistent 
cellular tree classifiers. Consider for example complete binary
trees with $k$ full levels, i.e., there are $2^k$ leaf regions.
We can have consistency
when $k$ is allowed to depend upon $n$. An example is the median
tree \citep[][Section 20.3]{DGL}. When $d = 1$, split by finding the median element 
among the $\bX_i$'s, so that the child sets have cardinality given by $\lfloor (n-1) /2\rfloor$  and $\lceil (n-1)/2\rceil$,
where $\lfloor . \rfloor$ and $\lceil.\rceil$ are the floor and ceiling functions. The median itself does stay behind and is not sent down to the subtrees, with an appropriate convention for breaking cell boundaries as well as empty cells. Keep doing this for $k$ rounds---in $d$ dimensions, one can either rotate through the coordinates for median splitting, or randomize by selecting uniformly at random a coordinate to split orthogonally.  
\medskip

This rule is known to be consistent as soon as the marginal distributions of $\bX$ are nonatomic, provided $k\to \infty$ and $k2^k/n \to 0$.  However, this 
is not a cellular tree classifier. While we can indeed specify $\sigma$, it is impossible to define $\theta$ because $\theta$ cannot
be a function of the global value of $n$. 
In other words, if we were to apply median splitting and decide to split for a fixed $k$, 
then the leaf nodes would all correspond to a fix proportion of the data points.
It is clear that the decisions in the leaves are off with a fair probability if we have, for example, $Y$ independent of $\bX$
 and $\mathbb P\{Y = 1\}=1/2$. Thus, we cannot create a cellular tree classifier in this manner.
\medskip

In view of the preceding discussion, it seems paradoxical that there indeed exist universally 
consistent cellular tree classifiers. 
(We note here that we abuse the word ``universal''---we will assume throughout,
to keep the discussion at a manageable level, that the marginal distributions
of $\bX$ are nonatomic. But no other conditions on the joint
distribution of $(\bX, Y)$ are imposed.) Our first construction, which is presented in Section 3, follows the median tree principle and uses randomization. In a second construction (Section 4) we derandomize, and exploit the idea that each cell is allowed to explore its own subtrees, thereby anticipating the decisions of its children. For the sake of clarity, proofs of the most technical results are gathered in Section 5 and Section 6.
\section{A randomized cellular tree classifier}
From now on, to keep things simple, it is assumed that the marginal distributions of $\bX$ are nonatomic.
The cellular splitting method $\sigma$ described in this section mimics the median tree classifier discussed above. 
We first choose a dimension to cut, uniformly at random
from the $d$ dimensions, as rotating through the dimensions by level number would
violate the cellular condition. The selected dimension is then split at the data median, just as in the classical median tree.
Repeating this for $k$ levels of nodes leads to $2^{k}$ leaf regions. On any path of length $k$ to one of the $2^{k}$ leaves, we have a deterministic sequence of cardinalities
$n_0=n (\mbox{root}), n_1, n_2, \hdots, n_k$. We always have $n_i/2-1\leq n_{i+1}\leq n_i/2$. Thus, by induction, one easily shows that, for all $i$,
$$
\frac{n}{2^i}-2\leq n_i \leq \frac{n}{2^i}.
$$
In particular, each leaf has at least $\max(n/2^{k}-2,0)$ points and at most $n/2^{k}$. 
\begin{rem}
\label{Remark3}
The problem of atoms in the coordinates can be dealt with separately, but still within
the cellular framework. 
The particularity is that the threshold for splitting may now be at a position at which one or
more data values occur. This leaves two sets that may differ in size by more than one. The atoms in the
distribution of $\bX$ can never be separated, but that is as it should be. We leave it to the reader to adapt the 
subsequent arguments to the case of atomic distributions.
\end{rem}
The novelty is in the choice of the decision function. This function ignores the data altogether
and uses a randomized decision that is based on the size of the input. 
More precisely, consider a nonincreasing function $\varphi:\mathbb N \to (0,1]$
with $\varphi(0)=\varphi(1)=1$. Cells correspond in a natural way to sets of $\mathbb R^d$.
So, we can and will speak of a cell $A$, where
$A \subset \mathbb R^d$.
The number of data points in $A$ is denoted by $N(A)$:
$$
N(A)= \sum_{i=1}^n \mathbf 1_{[\bX_i\in A]}.
$$
Then, if $U$ is the uniform $[0,1]$ random variable associated with the cell $A$
and the input to the cell is $N(A)$, the stopping rule \ding{172} takes the form:
\begin{center}
\fcolorbox{black}{gris25}{
\begin{minipage}{0.6\textwidth}
\begin{enumerate}
\item[\ding{172}] \textsf{Put $\theta=0$ if
$$U\leq \varphi\left(N(A)\right).$$}
\end{enumerate}
\end{minipage}
}
\end{center}
\medskip

In this manner, we obtain a possibly infinite randomized binary tree classifier. Splitting occurs
with probability $1-\varphi(n)$ on inputs of size $n$. 
Note that no attempt is made to split empty sets or singleton sets. 
For consistency, we need to look at the random
leaf region to which $\bX$ belongs. 
This is roughly equivalent to studying the distance from that cell to the
root of the tree.
\medskip

In the sequel, the notation $u_n=\mbox{o}(v_n)$ (respectively, $u_n=\omega(v_n)$ and $u_n=\mbox{O}(v_n)$) means that $u_n/v_n\to 0$ (respectively, $v_n/u_n\to 0$ and $u_n \leq C v_n$ for some constant $C$) as $n\to \infty$. Many choices $\varphi(n)=\mbox{o}(1)$, but not all, will do for us. The next lemma makes things more precise.
\begin{lem}
\label{J-8}
Let $\beta \in (0,1)$. Define 
\begin{eqnarray*} 
\varphi(n)=\left\{
\begin{array}{ll}
1 & \mbox{ if  $n<3$}\\
1/{\log^{\beta}n} & \mbox{ if  $n \geq 3$.}
\end{array}
\right.
\end{eqnarray*} 
Let $K(\bX)$ denote the random path distance between the cell of $\bX$ and the root of the tree. Then
\begin{eqnarray*} 
\lim_{n \to \infty} \mathbb P \left \{ K (\bX)\geq k_n\right\}=\left\{
\begin{array}{ll}
0 & \mbox{ if  $k_n=\omega(\log^{\beta} n)$}\\
1 & \mbox{ if  $k_n=\emph{o}(\log^{\beta} n)$.}
\end{array}
\right.
\end{eqnarray*} 
\end{lem}
\noindent{\bf Proof of Lemma \ref{J-8}}\quad Let us recall that, at level $k$, each cell  of the underlying median tree contains at least $\max(n/2^{k}-2,0)$ points and at most $n/2^{k}$. Since the function $\varphi(.)$ is nonincreasing, the first result follows from this:
\begin{align*}
\mathbb P \left \{ K(\bX) \geq k_n\right\} &\leq \prod_{i=0}^{k_n-1}\left (1-\varphi\left(\lfloor n/2^i\rfloor\right)\right)\\
& \leq \exp \left (- \sum_{i=0}^{k_n-1}\varphi\left (\lfloor n/2^i\rfloor\right) \right)\\
& \leq \exp\left (-k_n\varphi(n)\right).
\end{align*}
The second statement follows from
$$\mathbb P \left \{ K(\bX) < k_n\right\} \leq \sum_{i=0}^{k_n-1}\varphi\left (\lceil n/2^i-2\rceil \right)\leq k_n\varphi\left (\lceil n/2^{k_n}\rceil\right),$$
valid for all $n$ large enough since $n/2^{k_n}\to \infty$ as $n\to \infty$.
\hfill $\blacksquare$
\medskip

Lemma \ref{J-8}, combined with the median tree consistency result of \citet{DGL},
suffices to establish consistency of the randomized 
cellular  tree classifier. 
\begin{theo}
\label{theorem1}
Let $\beta$ be a real number in $(0,1)$. Define
\begin{eqnarray*} 
\varphi(n)=\left\{
\begin{array}{ll}
1 & \mbox{ if  $n<3$}\\
1/{\log^{\beta}n} & \mbox{ if  $n \geq 3$.}
\end{array}
\right.
\end{eqnarray*} 
Let $g_n$ be the associated randomized 
cellular  binary tree classifier. Assume that the marginal distributions of $\bX$ are nonatomic. Then
the classification rule $g_n$ is consistent:
$$\lim_{n \to \infty} \mathbb E L(g_n)=L^{\star}\quad \mbox{as } n \to \infty.$$
\end{theo}

\noindent{\bf Proof of Theorem \ref{theorem1}}\quad 
By $\mbox{diam}(A)$ we mean the diameter of the cell $A$, i.e., the maximal distance between two points of $A$. We recall a general consistency theorem for 
partitioning classifiers whose cell design depends on the $\bX_i$'s only \citep[][Theorem 6.1]{DGL}. According to this theorem, such a classifier is consistent if both
\begin{enumerate}
\item $\mbox{diam}(A(\bX)) \to 0 \quad \mbox{in probability as } n \to \infty,  \mbox{ and}$
\item $N(A(\bX))\to \infty\quad  \mbox{in probability as } n \to \infty,$
\end{enumerate}
where
$A(\bX)$ is the cell of the random partition
containing $\bX$.
\medskip

Condition 2.~is proved in Lemma \ref{J-8}. 
Notice that
\begin{align*}
N\left(A(\bX)\right) &\geq \frac{n}{2^{K(\bX)}} - 2 \\
& \geq \mathbf 1_{[K(\bX) < \log^{(\beta+1)/2}n]} \left( \frac{n}{2^{\log^{(\beta+1)/2} n }} - 2 \right) \\
&= \omega(1)\mathbf 1_{[K(\bX) < \log^{(\beta+1)/2}n]}.
\end{align*}
Therefore, by Lemma \ref{J-8}, $N\left(A(\bX)\right) \to \infty$ in probability as $n \to \infty$.
\medskip

To show that $\mbox{diam}(A(\bX)) \to 0$ in probability, observe that on a path of length
$K(\bX)$, the number of times the first dimension is cut is binomial $(K(\bX),1/d)$. 
This tends to infinity in probability.
Following the proof of Theorem 20.2 in \citet{DGL}, 
 the diameter of the cell of $\bX$ tends to $0$ in probability with $n$. Details are left to the reader.
\hfill $\blacksquare$
\medskip

Let us finally take care of the randomization. Can one do without randomization? The hint to the
solution of that enigma is in the hypothesis that the data elements in $\mathcal D_n$ are i.i.d. The median classifier
does not use the ordering in the data. Thus, one can use the randomness present in the permutation
of the observations, e.g., the $\ell$-th components of the $\bX_i$'s can form $n!$ permutations if ties do not occur. This
corresponds to $(1+\mbox{o}(1))n \log_2 n$ independent fair coin flips, which are at our disposal. Each decision to
split requires on average at most $2$ independent bits. The selection of a random direction to cut requires
no more than $1+\log_2 d$ independent bits. Since the total tree size is, with probability tending to 1, $\mbox{O}(2^{\log^{\beta+\varepsilon} n})$ for any $\varepsilon >0$, a fact that follows 
with a bit of work from summing the expected number of nodes at each level,
the total number of bits required to carry out all computations is
$$\mbox{O} \left ( (3+\log_2 d)2^{\log^{\beta+\varepsilon} n}\right),$$
which is orders of magnitude smaller than $n$
provided that $\beta + \varepsilon < 1$. 
Thus, there is sufficient randomness at hand to do the job.
How it is actually implemented is another matter, as there is some inevitable dependence between the
data sets that correspond to cells and the data sets that correspond to their children. We will not worry
about the finer details of this in the present paper. 
\begin{rem}
\label{Remark4}
For more on random tree models and their analyses, see
the texts of \citet{Drm09}, and \citet{FlaSed08}.
Additional
material on information-theory and bit complexity
can be found in the monograph by \citet{Coto91}.
\end{rem}
\begin{rem}
\label{Remark++}
In the spirit of Breiman's random forests \citeyearpar{Bre01}, one could envisage to use a collection of randomized cellular tree classifiers and make final predictions by aggregating over the ensemble. Since each individual rule is consistent (by Theorem \ref{theorem1}), then the same property is also true for the ensemble \citep[see, e.g., Proposition 1 in][]{BDL08}. Improvements are expected at the level of predictive accuracy and stability.
\end{rem}
\section{A non-randomized cellular tree classifier}
The cellular tree classifier that we consider in this section is more sophisticated and autonomous,
in the sense that it does not rely on any randomization scheme. It partitions the data recursively as follows. 
With each node we associate a set of $\mathbb R^d$, starting with $\mathbb R^d$ for the root node. 
We first consider a full $2^d$-ary tree (see Figure \ref{figure3} for an illustration in dimension 2), with the cuts
decided in the following manner. The dimensions are ordered once and for all
from $1$ to $d$. At the root, we find the median of (the projection of) the $n$ data points in direction $1$,
then on each of the two subsets, we find the median in direction $2$,
then on each of the four subsets, we find the median in direction $3$,
and so forth. 
A split, contrary to our discussion thus far, is into $2^d$ parts, not two parts.
This corresponds to Bentley's $k$-$d$ tree \citeyearpar{Bentley75}.
Repeating this splitting
for $k$ levels of nodes leads to $2^{dk}$ leaf regions, each having at least $\max(n/2^{dk}-2,0)$ points and at most $n/2^{dk}$.
\begin{figure}[h]
\centering
\psset{treemode=D,treesep=10pt,levelsep=35pt}

\pstree[treemode=D,treesep=10pt,levelsep=35pt]{\Tc{1.5mm}}
{

\pstree{\Tc{1.5mm}}
{

\pstree{\Tc{1.5mm}}{\psset{linestyle=dotted}\pstree{\Tp}{}}
\pstree{\Tc{1.5mm}}{\psset{linestyle=dotted}\pstree{\Tp}{}}
\pstree{\Tc{1.5mm}}{\psset{linestyle=dotted}\pstree{\Tp}{}}
\pstree{\Tc{1.5mm}}{\psset{linestyle=dotted}\pstree{\Tp}{}}
}

\pstree{\Tc{1.5mm}}
{

\pstree{\Tc{1.5mm}}{\psset{linestyle=dotted}\pstree{\Tp}{}}
\pstree{\Tc{1.5mm}}{\psset{linestyle=dotted}\pstree{\Tp}{}}
\pstree{\Tc{1.5mm}}{\psset{linestyle=dotted}\pstree{\Tp}{}}
\pstree{\Tc{1.5mm}}{\psset{linestyle=dotted}\pstree{\Tp}{}}
}

\pstree{\Tc{1.5mm}}
{

\pstree{\Tc{1.5mm}}{\psset{linestyle=dotted}\pstree{\Tp}{}}
\pstree{\Tc{1.5mm}}{\psset{linestyle=dotted}\pstree{\Tp}{}}
\pstree{\Tc{1.5mm}}{\psset{linestyle=dotted}\pstree{\Tp}{}}
\pstree{\Tc{1.5mm}}{\psset{linestyle=dotted}\pstree{\Tp}{}}
}

\pstree{\Tc{1.5mm}}
{

\pstree{\Tc{1.5mm}}{\psset{linestyle=dotted}\pstree{\Tp}{}}
\pstree{\Tc{1.5mm}}{\psset{linestyle=dotted}\pstree{\Tp}{}}
\pstree{\Tc{1.5mm}}{\psset{linestyle=dotted}\pstree{\Tp}{}}
\pstree{\Tc{1.5mm}}{\psset{linestyle=dotted}\pstree{\Tp}{}}
}

}
\psset{linestyle=dotted}
\hspace{0.5cm}
\pstree[treemode=D,treesep=10pt,levelsep=35pt]{\Tr{$k=0$}}
{

\pstree{\Tr{$k=1$}}{

\pstree{
\Tr{$k=2$}} 
{\psset{linestyle=dotted}\pstree{\Tp}{}}

}
}
\psset{linestyle=solid}
\caption{\label{figure3}\textsf{A full $2^d$-ary tree in dimension $d=2$. }}
\end{figure}

This procedure is equivalent to $dk$ consecutive binary splits at the
median, where we rotate through the dimensions. However, in our 
cellular set-up, such rotations through the dimensions are
impossible, and this forces us to employ this equivalent strategy. 
Note, therefore, that the
split parameter $\sigma$ is an extension
of the binary classifier split $\sigma$---one could consider
it as a vector of dimension $2^d -1$, as we need to specify $2^d -1$
coordinate positions to fully specify a partition into $2^d$ regions. 
It remains to specify a stopping rule $\theta$ which 
respects the cellular constraint. To this aim, we need some additional notation.
\begin{rem}
\label{Remark5}
By the very construction of the tree, at each node, the median itself does stay behind and is not sent down to the subtrees. From a topological point of view, this means that, in the partition building, each cell $A$ and its $2^d$ child cells $A_1, \hdots, A_{2^d}$ are considered as {\bf open} hyperrectangles. Thus, for classification, assuming nonatomic marginals, we would thus strictly speaking not be able to classify any data that fall ``on the border'' between $A_1, \hdots, A_{2^d}$. This is a non-important detail for the calculations since the marginal distributions of $\bX$ are nonatomic. In practice, this issue can be solved with an appropriate convention to break the boundary ties.
\end{rem}
If $A$ is any cell of the full $2^d$-ary tree defined above, we let $N(A)$ be the number of $\bX_i$'s falling
in $A$, and estimate the quality of the majority vote classifier
at this node by
$$\hat L_n(A)=\frac{1}{N(A)}\min \left (\sum_{i=1}^n \mathbf 1_{[\bX_i \in A, Y_i=1]}, \sum_{i=1}^n \mathbf 1_{[\bX_i \in A, Y_i=0]}\right).$$
(Throughout, we adopt the convention $0/0=0$.) 
\begin{rem}
\label{Remark6}
Each cut at the median eliminates 1 data point. Thus, given a cell $A$, the construction of its offspring $k$ generations later rules out at most $1+\cdots+2^{dk-1}=2^{dk}-1$ observations. In particular, if $A$ has cardinality $N(A)$, then, $k$ generations  later, its offspring $A_1, \hdots, A_{2^{dk}}$ have a total combined
cardinality at least $N(A)-(2^{dk}+1)$.
\end{rem}
Fix a positive real parameter $\alpha$ and define the nonnegative integer $k^+$ by
$$
k^+= \left \lfloor {\alpha \log_2  (N(A)+1) } \right \rfloor,
$$
where, for simplicity, we drop the dependency of $k^+$ upon $A$ and $\alpha$.
Finally, letting $\mathcal P_{k^+}(A)$ be the $2^{dk^+}$ leaf regions (terminal nodes) of the 
full $2^d$-ary tree rooted at $A$ of height $k^+$, we set
$$\hat L_{n}(A,k^+)=\sum_{A_j \in \mathcal P_{k^+}(A)}\hat L_n(A_j)\frac{N(A_j)}{N(A)}.$$
The quantity $\hat L_{n}(A,k^+)$ is interpreted as the total (normalized) error of a majority
vote over the offspring of $A$ living $k^+$ generations later. It should be stressed that {\bf both} $\hat L_{n}(A)$
and  $\hat L_{n}(A,k^+)$ may be evaluated on the basis of the data points falling in $A$ only (no matter what the rest of the tree looks like),
thereby respecting the cellular constraint.
\medskip

Now, let $\beta$ be a positive real parameter. With this notation, the stopping rule \ding{172} takes the following simple form:
\begin{center}
\fcolorbox{black}{gris25}{
\begin{minipage}{0.6\textwidth}
\begin{enumerate}
\item[\ding{172}] \textsf{Put $\theta=0$ if
$$\left |\hat L_n(A)- \hat L_{n}(A,k^+)\right|\leq \left(\frac{1}{N(A)+1}\right)^{\beta}.$$}
\end{enumerate}
\end{minipage}
}
\end{center}
\medskip

In other words, at each cell, the algorithm compares the actual classification error with the total error of the cell offspring $k^+$ generations later. 
This bounded lookahead principle suggested by us 
is quite well-developed in the artificial intelligence
literature---see, for example, Pearl's book \citeyearpar{Pearl88} on
probabilistic reasoning.
If the difference is below some well-chosen threshold, then the cellular
classification procedure stops and the node returns a terminal signal. Otherwise, the node outputs $2^{d}$ sets of data, and the process continues recursively. 
The protocol stops once all nodes have returned a terminal signal, and final decisions are taken by majority vote. Thus, for $\bx$ falling in a terminal node $A$, the rule is as usual
\begin{eqnarray*} 
g_n (\bx) = \left\{
\begin{array}{ll}
1 & \mbox{ if  $\sum_{i=1}^n \mathbf 1_{[\bX_i \in A, Y_i =1]} > \sum_{i=1}^n \mathbf 1_{[\bX_i \in A, Y_i =0]}$}\\
0 & \mbox{ otherwise.}
\end{array}
\right.
\end{eqnarray*} 

In the next section, we prove the following theorem:
\begin{theo}
\label{convergence}
Let $g_n$ be the cellular tree classifier defined above, with $1-d\alpha -2\beta >0$.
Assume that the marginal distributions of $\bX$ are nonatomic. Then
the classification rule $g_n$ is consistent:
$$\lim_{n \to \infty} \mathbb E L(g_n)=L^{\star}\quad \mbox{as } n \to \infty.$$
\end{theo}

From a technical point of view, this theorem poses a challenge, as there
are no conditions on the distribution, and the rectangular cells do in
general not shrink to zero. In fact, it is easy to find distributions
of $\bX$ for which the maximal cell diameter does not tend to zero
in probability, even if all is restricted to the unit cube. For distributions
with infinite support, there are always cells of infinite diameter.
This observation implies that classical consistency proofs, that often use
differentiation of measure arguments or rely on asymptotic justifications
related to Lebesgue's density theorem, cannot be applied.
The proof uses global arguments instead.
\medskip

For partitions that do not depend upon the $Y$-values in the data,
consistency can be shown by relatively simple means, following for example the
arguments given in \citet{DGL}. However, our partition and
tree depend upon the $Y$-values in the data. Within the constraints
imposed by the cellular model, we believe that this is the first (and only)
proof of universal consistency of a $Y$-dependent cellular tree classifier.
On the other hand, we have proposed a model that is a priori too simple to
be competitive. There are choices of parameters to be made,
and there is absolutely no minimax theory of lower bounds for
the rate with which cellular tree classifiers can
approach the Bayes error. On the practical side, besides the question of how to 
efficiently implement the model, it is also clear that 
the performance of the cellular estimate will be conditional on a good tuning
of both parameters $\alpha$ and $\beta$. As a first step, a good route
to follow is to attack the rate of convergence problem---we expect dependence on the smoothness of $(\bX,Y)$---and deduce from this analysis the best parameter choices. In any case, the work ahead is enormous
and the road arduous.
\section{Proof of Theorem \ref{convergence}}
\subsection{Notation and preliminary results}
We start with some notation (see Figure \ref{figure4}). For each level $k\geq 0$, we denote by $\mathcal P_{k}$ the partition represented by the leaves of the underlying full $2^{d}$-ary median-type tree.  
This partition has $2^{dk}$ cells and its construction depends on the $\bX_i$'s only. The labels $Y_i$'s do not play a role in the building of $\mathcal P_{k}$, though they are involved in making the decision whether to cut a cell or not. 
\begin{figure}[!t]
\vspace{4cm}
\begin{minipage}[b]{.75\linewidth}
\centering
\begin{pspicture}[unit=2.2cm](0,0)(5,6)
\rput(2,4.15){\psframebox{root}}
\psline(0,0)(2,4)(4,0)
\psline[linewidth=2pt](1,2)(3,2)
\psset{linestyle=dotted}
\psline(0,0)(-0.1,-0.2)
\psline(4,0)(4.1,-0.2)
\psset{linestyle=solid}
\pscurve[linewidth=2pt](0.5,1)(1.1,1.1)(1.5,2.5)(1.8,1.7)(2,1.7)(2.25,2.9)(2.8,0.7)(3,0.6)(3.2,1.6)
\psset{arrowscale=2,arrows=<->,linestyle=dotted}
\psline(3,2)(3,4)
\psset{arrowscale=2,arrows=<-,linestyle=dotted}
\psline(3,2)(4,2)
\uput[u](3.2,2.8){$k$}
\uput[u](4.2,1.81){$\mathcal P_{k}$}
\psset{arrowscale=2,arrows=-,linestyle=solid}
\pscircle*(1.5,1.3){0.05}
\psline(1.5,1.3)(1,0.3)
\psline(1.5,1.3)(2,0.3)
\psline[linewidth=2pt](1,0.3)(2,0.3)
\uput[u](1.5,1.3){$A$}
\psset{arrowscale=2,arrows=<->,linestyle=dotted}
\psline(2,0.3)(2,1.3)
\uput[u](2.2,0.6){$k^+$}
\uput[u](1.5,-0.1){$\mathcal P_{k^+}(A)$}
\psset{arrowscale=2,arrows=->,linestyle=dotted}
\psline(0.2,2.8)(1.1,1.12)
\psline(0.2,2.8)(1.28,2.3)
\uput[u](0.2,2.8){$\mathcal G_n$}
\psline(4,1.1)(2.65,1.1)
\uput[u](4.2,0.9){$\mathcal G^+_{k}$}
\psline(1,3.3)(2.08,2.6)
\uput[u](1,3.3){$\mathcal G^-_{k}$}
\end{pspicture}
\end{minipage}
\caption{\label{figure4}\textsf{Some key notation.}}
\end{figure}

For each $A_j \in \mathcal P_{k}$, we let $N(A_j)$ be the number of $\bX_i$'s falling in $A_j$ and note that $\sum_{j=1}^{2^{dk}}N(A_j)\leq n$, with a strict inequality as soon as $k>0$ (see Remark \ref{Remark6}). For each level $k$, $A_k(\bX)$ denotes the cell of the partition $\mathcal P_{k}$ into which $\bX$ falls, and $N(A_k(\bX))$ the number of data points falling in this set. 
 \medskip
 
We let $\mu$ be the distribution of $\bX$ and $\eta$ the regression function of $Y$ on $\bX$. More precisely, for any Borel-measurable set $A \subset \mathbb R^d$,
 $$\mu(A)=\mathbb P\{\bX \in A\}$$
 and, for any $\bx \in \mathbb R^d$,
 $$\eta(\bx)=\mathbb P\{Y=1|\bX=\bx\}=\mathbb E [Y | \bX=\bx].$$
It is known that the Bayes error is
$$L^{\star}=\int_{\mathbb R^d} \min \left (\eta(\bz),1-\eta(\bz)\right)\mu(\mbox{d} \bz).$$
Let us recall that, for any cell $A$,
$$\hat L_n(A)=\frac{1}{N(A)}\min \left (\sum_{i=1}^n \mathbf 1_{[\bX_i \in A, Y_i=1]}, \sum_{i=1}^n \mathbf 1_{[\bX_i \in A, Y_i=0]}\right).$$
Also, for every $k\geq 0$,
$$\hat L_{n}(A,k)=\sum_{A_j \in \mathcal P_{k}(A)}\hat L_n(A_j)\frac{N(A_j)}{N(A)},$$
where $\mathcal P_{k}(A)$ is the full $2^d$-ary median-type tree rooted at $A$ of height $k$. 
At the population level, we set 
$$L^{\star}(A)= \frac{1}{\mu(A)}\min \left (\int_{A}\eta(\bz)\mu(\mbox{d}\bz), \int_{A}\left (1-\eta(\bz)\right)\mu(\mbox{d}\bz)\right)$$
and  
$$L^{\star}(A,k)Ê = \sum_{A_j \in \mathcal P_{k}(A)} L^{\star}(A_j) \frac{\mu(A_j)}{\mu(A)}.$$
For all $k\geq 0$, we shall also need the quantity
$$L^{\star}_{k}=\mathbb E \left [ÊL^{\star}\left(A_k(\bX)\right)\right].$$
Note that whenever $A=A(\bX_1, \hdots, \bX_n)$ is a random cell, we take the liberty to abbreviate
$\int_A\mbox{d}\mu$ by $\mu(A)$ throughout the manuscript, 
since this should cause no confusion. We write for instance
$$L^{\star}_k=\mathbb E \big [\mathbb E \left [L^{\star}\left(A_k(\bX)\right)\,|\,\bX_1, \hdots, \bX_n\right]\big]=\mathbb E \left [ \sum_{A \in \mathcal P_k} L^{\star}(A)\mu(A)\right]$$
instead of 
$$L^{\star}_{k} =\mathbb E \left [ \sum_{A \in \mathcal P_k} L^{\star}(A)\int_A\mbox{d}\mu\right].$$
\medskip

Our proof starts with some easy but important facts.
 \begin{fact}
 \label{fait1}
 \begin{enumerate}
 \item[]
 \item[$(i)$] For all levels $k'\geq k \geq 0$,
 $$L^{\star}\leq L^{\star}_{k'}\leq L^{\star}_{k}.$$
 \item[$(ii)$] For each cell $A$ and each level $k\geq 0$,
$$\hat L_n(A,k) \leq \hat L_n(A)+\frac{2^{dk}}{N(A)}\mathbf 1_{[N(A)>0]}.$$
 \item[$(iii)$] For each cell $A$ and all levels $k'\geq k \geq 0$,
$$\hat L_n(A,k')\leq \hat L_n(A,k)+\frac{2^{dk'}}{N(A)}\mathbf 1_{[N(A)>0]}.$$
\item[$(iv)$] For each cell $A$ and all levels $k,k'\geq0$,
$$\mathbb E \left [L^{\star}(A_k(\bX),k')\right]Ê=L^{\star}_{k+k'}.$$
In particular, for $k''\geq k' \geq 0$,
$$L^{\star}\leq \mathbb E \left [ L^{\star}(A_k(\bX),k'')\right]\leq \mathbb E \left [ L^{\star}(A_k(\bX),k')\right].$$
 \end{enumerate}
 \end{fact}
\noindent{\bf Proof}\quad Proof of statement $(i)$ is
based on the nesting of the partitions. To establish $(ii)$, observe that, by definition,
$$\hat L_n(A)=\frac{1}{2}-\frac{1}{2N(A)}\left | N(A) -2 \sum_{i=1}^n \mathbf 1_{[\bX_i \in A, Y_i=1]}\right|,$$
and
\begin{align*}
&\hat L_n(A,k) \\
& \quad =\frac{1}{2N(A)}\sum_{A_j \in \mathcal P_k(A)}N(A_j)-\frac{1}{2N(A)}\sum_{A_j \in \mathcal P_k(A)} \left |N(A_j)-2 \sum_{i=1}^n \mathbf 1_{[\bX_i \in A_j,Y_i=1]}\right|\\
&\quad \leq \frac{1}{2}-\frac{1}{2N(A)}\sum_{A_j \in \mathcal P_k(A)} \left |N(A_j)-2 \sum_{i=1}^n \mathbf 1_{[\bX_i \in A_j,Y_i=1]}\right|.
\end{align*}
But, by the triangle inequality and Remark \ref{Remark6},
\begin{align*}
&\left |N(A)-2\sum_{i=1}^n \mathbf 1_{[\bX_i \in A,Y_i=1]}\right| \\
& \quad \leq \sum_{A_j \in \mathcal P_k(A)} \left |N(A_j)-2 \sum_{i=1}^n \mathbf 1_{[\bX_i \in A_j,Y_i=1]}\right| + 2^{dk}-1.
\end{align*}
This proves $(ii)$. Proof of $(iii)$ is similar. To show $(iv)$, just note that
 \begin{align*}
\mathbb E \left [L^{\star}(A_k(\bX),k')\right]Ê & = \mathbb E \left [\sum_{A \in \mathcal P_{k}} \sum_{A_j \in \mathcal P_{k'}(A)} L^{\star}(A_j) \frac{\mu(A_j)}{\mu(A)}\mu(A)\right]\\
& = \mathbb E \left [ L^{\star}\left (A_{k+k'}(\bX)\right)\right]\\
& =  L^{\star}_{k+k'}.
\end{align*}
\hfill $\blacksquare$
\medskip 

The next two propositions will be decisive in our analysis. Proposition \ref{diametre} asserts that the diameter of $A_k(\bX)$ tends to 0 in probability, provided $k$ (as a function of $n$) tends sufficiently slowly to infinity.  Proposition \ref{lemmepsi} introduces a particular level $k_n^{\star}$ which will play a central role in the proof of Theorem \ref{convergence}.
 \begin{pro}
 \label{diametre}
 Assume that the marginal distributions of $\bX$ are non\-ato\-mic. Then, if
 $$k \to \infty \quad \mbox{and} \quad \frac{k2^{dk}}{n} \to 0,$$
 one has
 $$\emph{diam}\left (A_k(\bX)\right)\to 0 \quad \mbox{in probability as } n \to \infty.$$
 \end{pro}
 \noindent{\bf Proof of Proposition \ref{diametre}}\quad 
 Median-split trees are analyzed in some detail in Section 20.3 of the monograph by \citet{DGL}. 
 Starting on page 323, it is shown that the diameter of a randomly selected cell tends to 0 in probability. The adaptation to our $2^d$-ary median-type trees is straightforward.
 However, a few remarks are in order. ÊSection 20.3 of that book assumes that all marginals are
uniform. ÊThis can also be the set-up for us, because our rule
is invariant under monotone transformations of the axes.
Note however that it is crucial that splits are made exactly at data points for this
property to be true. Also, the proofs in Section 20.3 of \citet{DGL} assume $d=2$, but are clearly true
for general $d$. The only condition for the diameter result is that of Theorem 20.2, page 323:
$$k \to \infty \quad \mbox{and} \quad \frac{k2^{dk}}{n}\to \infty.$$
The second condition is only necessary to make sure that the data medians do
not run too far away from the true distributional medians. Ê
 \hfill $\blacksquare$
\medskip
\begin{pro}
\label{lemmepsi}
Let $\psi(n,k)$ be the function defined for all $n \geq 1$ and $k \geq 0$ by
$$\psi(n,k)=L_{k}^{\star}-L^{\star}.$$
\begin{enumerate}
\item[$(i)$] Let $\{k_n\}_{n \geq 1}$ be a sequence of nonnegative integers such that $k_n\to \infty$ and $k_n2^{dk_n}/n \to 0$. Then
$$\psi(n,k_n)\to 0\quad \mbox{as } n \to \infty.$$
\item[$(ii)$] Assume that $\alpha \in (0,1/d)$ and, for fixed $n$, set 
$$k_n^{\star}=\min \left \{ \ell \geq 0: \psi(n,\ell)<\sqrt {\left(\frac{ 2^{d\ell }}{n}\right)^{1-d\alpha}}\right\}.$$
Then
$$ \frac{2^{dk^{\star}_n}}{n} \to 0 \quad \mbox{as } n \to \infty.$$
\end{enumerate}
\end{pro}
\noindent{\bf Proof of Proposition \ref{lemmepsi}} \quad 
At first we note, according to Fact \ref{fait1}$(ii)$, that for all $n\geq 1$ and $k\geq 0$, $\psi(n,k)\geq 0$. For $\bx \in \mathbb R^d$, introduce
$$\bar \eta_{n}(\bx)=\frac{1}{\mu\left(A_{k_n}(\bx)\right)}\int_{A_{k_n}(\bx)}\eta(\bz)\mu(\mbox{d}\bz).$$
With this notation,
\begin{align*}
\psi(n,k)& = \mathbb E \left [L^{\star}\left(A_{k_n}(\bX)\right)\right]-L^{\star}\\
& \leq \mathbb E \left |\eta(\bX)-\bar \eta_{n}(\bX) \right|+\mathbb E\left |\left(1-\eta(\bX)\right)-\left(1-\bar \eta_{n}(\bX)\right)\right|.
\end{align*}
Let us prove that the first of the two terms above tends to 0 as $n$ tends to infinity---the second term is handled similarly. To this aim, fix $\varepsilon>0$ and find a uniformly continuous function $\eta_{\varepsilon}$ on a bounded set $\mathcal C$ and vanishing off $\mathcal C$ so that $\mathbb E |\eta(\bX)-\eta_{\varepsilon}(\bX)|<\varepsilon$. Clearly, by the triangle inequality,
\begin{align*}
 \mathbb E \left |\eta(\bX)-\bar \eta_{n}(\bX) \right| &\leq \mathbb E \left |\eta(\bX)-\eta_{\varepsilon}(\bX)\right|\\
 & \quad + \mathbb E \left |\eta_{\varepsilon}(\bX)-\bar \eta_{n,\varepsilon}(\bX)\right|\\
 & \quad + \mathbb E \left |\bar \eta_{n,\varepsilon}(\bX)-\bar \eta_{n}(\bX)\right|\\
&  \stackrel{\mbox{\footnotesize def}}{=}\mbox{\sc I}+\mbox{\sc II}+ \mbox{\sc III},
\end{align*}
where
$$\bar \eta_{n,\varepsilon}(\bx)=\frac{1}{\mu\left(A_{k_n}(\bx)\right)}\int_{A_{k_n}(\bx)}\eta_{\varepsilon}(\bz)\mu(\mbox{d}\bz).$$
By choice of $\eta_{\varepsilon}$, one has $\mbox{\sc I}<\varepsilon$. Next, note that
$$\mbox{\sc II} \leq \mathbb E \left [\frac{\displaystyle \int_{A_{k_n}(\bX)} \left | \eta_{\varepsilon}(\bX)-\eta_{\varepsilon}(\bz)\right|\mu(\mbox{d}\bz)}{\mu\left (A_{k_n}(\bX)\right)}\right].$$
As $\eta_{\varepsilon}$ is uniformly continuous, there exists a number $\delta=\delta(\varepsilon)>0$ such that if $\mbox{diam}(A)\leq \delta$, then $|\eta_{\varepsilon}(\bx)-\eta_{\varepsilon}(\bz)|<\varepsilon$ for every $\bx, \bz\in A$. In addition, there is a positive constant $M$ such that $|\eta_{\varepsilon}(\bx)|\leq M$ for every $\bx \in \mathbb R^d$. 
Thus,
$$\mbox{\sc II}<\varepsilon+2M\,\mathbb P \left \{\mbox{diam}\left(A_{k_n}(\bX)\right)>\delta\right\}.$$
Therefore, $\mbox{\sc II}<2\varepsilon$ for all $n$ large enough by Proposition \ref{diametre}. Finally, 
$\mbox{\sc III}\leq\mbox{\sc I}<\varepsilon$. Taken together, these steps prove the first statement of the proposition.
\medskip

Next, suppose assertion $(ii)$ is false and set, to simplify notation, $\delta=1-d\alpha>0$. Then we can find a subsequence $\{k^{\star}_{n_i}\}_{i\geq 1}$ of $\{k^{\star}_n\}_{n \geq 1}$ and a positive constant $C$ such that, for all $i$,
$$\frac{2^{dk^{\star}_{n_i}}}{n_i} \geq C.$$
Since $n_i \to \infty$, it can be assumed, without loss of generality, that $n_i\geq 2$
and $\log_2(Cn_i)\geq 2d$ for all $i$. This implies in particular
\begin{align}
k^{\star}_{n_i} -1&\geq \frac{\log_2(C n_i)}{d}-1 \nonumber\\
& \geq \frac{\log_2(Cn_i)}{2d},\label{10:02}
\end{align}
and $k^{\star}_{n_i} \geq 2$ as well.
\medskip

On the one hand, by the very definition of $k^{\star}_{n_i}$,
\begin{align}
\psi(n_i,k^{\star}_{n_i}-1)&\geq\sqrt{\left(\frac{2^{d(k^{\star}_{n_i}-1)}}{n_i}\right)^{\delta}} \nonumber\\
&\geq \sqrt{\frac{C^{\delta}}{2^{d\delta}}}\label{borneinf}.
\end{align}
On the other hand, by (\ref{10:02}) and the monotonicity of $\psi(n_i,.)$ (Fact \ref{fait1}$(ii)$), we may write
$$\psi(n_i,k^{\star}_{n_i}-1)\leq \psi \left(n_i,\frac{\log_2 (C n_i)}{2d}\right).$$
But, setting
$$t_{n_i}=\frac{\log_2 (C n_i)}{2d},$$
we have
$$\frac{t_{n_i}2^{dt_{n_i}}}{n_i}= \frac{\log_2(Cn_i)}{2d}\sqrt{\frac{C}{n_i}}.$$
This quantity goes to $0$ as $n_i \to \infty$. Moreover, $t_{n_i}\to \infty$ and thus, according to the first statement of the proposition, 
$$\psi(n_i,k^{\star}_{n_i}-1) \to 0 \quad \mbox{as } n_i \to \infty.$$
This contradicts (\ref{borneinf}).\hfill $\blacksquare$
\subsection{Proof of the theorem}
Let $\left \{k^{\star}_n\right\}_{n \geq 1}$ be defined as in Proposition \ref{lemmepsi}. We denote by $\mathcal G_n$ the leaf regions of the cellular tree, and by $\mathcal G_{k_n^{\star}}^-$ (respectively, $\mathcal G_{k_n^{\star}}^+$) the collection of leaves at level at most (respectively, strictly at least) $k^{\star}_n$. Finally, for any cell $A$, we set
$$L_n(A)=\mathbb P\{g_n(\bX)\neq Y, \bX\in A\,|\,\mathcal D_n\}.$$
With this notation, we have
\begin{align*}
L^{\star} \leq \mathbb EL(g_n)&=\mathbb E \left [\sum_{A \in \mathcal G_n} L_n(A)\right]\\
& =\mathbb E\left [ \sum_{A \in \mathcal G^-_{k^{\star}_n}} L_n(A)\right]+\mathbb E\left [\sum_{A \in \mathcal G^+_{k^{\star}_n}} L_n(A)\right].
\end{align*}
Set
$$\varphi(A)=\left(\frac{1}{N(A)+1}\right)^{\beta}.$$
Then, clearly, 
\begin{align*}
\mathbb E \left [\sum_{A \in \mathcal G^+_{k^{\star}_n}} L_n(A)\right]& \leq \mathbb E \left [\sum_{A \in \mathcal G^+_{k^{\star}_n}} \mu(A)\right]\\
& \leq \mathbb E \left [\sum_{A \in \mathcal P_{k^{\star}_n}}\mathbf 1_{[|\hat L_n(A)-\hat L_n(A,k^+)|>\varphi(A)]}Ê\mu(A)\right]\\
& =  \mathbb P  \left \{ \left |\hat L_{n}\left(A_{k_n^{\star}}(\bX)\right) -\hat L_{n}(A_{k^{\star}_n}(\bX),k^+)\right|>\varphi\left (A_{k_n^{\star}}(\bX)\right)\right\}.
\end{align*}
In the second inequality, we used the definition of the stopping rule of the cellular tree. Therefore, according to technical Lemma \ref{greedymoins}, 
$$\mathbb E \left [\sum_{A \in \mathcal G^+_{k^{\star}_n}} L_n(A)\right]\leq \mbox{O} \left (\sqrt{\left (\frac{2^{dk_n^{\star}}}{n}\right)^{1-d\alpha-2\beta}}\right).$$
Since $1-d\alpha-2\beta>0$, this term tends to 0 as $n \to \infty$ by the second statement of Proposition \ref{lemmepsi}. Next,  introduce the notation
$$N_0(A)=\sum_{i=1}^n \mathbf 1_{[\bX_i \in A, Y_i=0]} \quad \mbox{and} \quad N_1(A)=\sum_{i=1}^n \mathbf 1_{[\bX_i \in A, Y_i=1]},$$
and observe that
\begin{align*}
\mathbb E\left [ \sum_{A \in \mathcal G^-_{k^{\star}_n}} L_n(A)\right]&=\mathbb E \Bigg [\sum_{A \in \mathcal G^-_{k^{\star}_n}} \bigg\{\mathbf 1_{[N_0(A)\geq N_1(A)]} \int_A \eta(\bz)\mu(\mbox{d}\bz)\\
& \quad \qquad +\mathbf 1_{[N_0(A)< N_1(A)]} \int_A \left( 1-\eta(\bz)\right)\mu(\mbox{d}\bz)\bigg\}\Bigg].
\end{align*}
For $\bx$ falling in the region covered by $\mathcal G^-_{k^{\star}_n}$, denote by $A_{k_n^{\star}}^-(\bx)$ the cell of $\mathcal G^-_{k^{\star}_n}$ containing $\bx$, and set 
$$N(A_{k_n^{\star}}^-(\bx))=\sum_{i=1}^n \mathbf 1_{[\bX_i \in A_{k_n^{\star}}^-(\bx)]}.$$ 
Letting
$$\hat \eta_n(\bx)=\frac{1}{N\big(A_{k_n^{\star}}^-(\bx)\big)}\sum_{i=1}^n \mathbf 1_{[\bX_i \in A_{k_n^{\star}}^-(\bx),Y_i=1]},$$
we may write
\begin{align*}
&\mathbb E\left [ \sum_{A \in \mathcal G^-_{k^{\star}_n}} L_n(A)\right]\\
& \quad  \leq \mathbb E \Bigg [\sum_{A \in \mathcal G^-_{k^{\star}_n}} \hat L_n(A)\mu(A)\\
& \quad + \sum_{A \in \mathcal G^-_{k^{\star}_n}}\left \{\mathbf 1_{[N_0(A)\geq N_1(A)]} \left (\int_A \eta(\bz)\mu(\mbox{d}\bz)- \int_A \hat \eta_n(\bz)\mu(\mbox{d}\bz)\right)\right\}\\
& \quad + \sum_{A \in \mathcal G^-_{k^{\star}_n}}\left \{\mathbf 1_{[N_0(A)< N_1(A)]} \left (\int_A \left(1-\eta(\bz)\right)\mu(\mbox{d}\bz)- \int_A \left (1-\hat \eta_n(\bz)\right)\mu(\mbox{d}\bz)\right)\right\}\Bigg].
\end{align*}
It follows, evoking Lemma \ref{greedyplus}, that
$$\mathbb E\left [ \sum_{A \in \mathcal G^-_{k^{\star}_n}}L_n(A)\right] \leq \mathbb E \left [\sum_{A \in \mathcal G^-_{k^{\star}_n}} \hat L_n(A)\mu(A)\right]+\mbox{O}\left ( \sqrt{\frac{2^{dk^{\star}_n}}{n}}\right).$$
The rightmost term tends to 0 according to the second statement of Proposition \ref{lemmepsi}. 
\medskip

Thus, to complete the proof, it remains to establish that
$$\mathbb E \left [\sum_{A \in \mathcal G^-_{k^{\star}_n}} \hat L_n(A)\mu(A)\right] \to L^{\star} \quad \mbox{as }n \to \infty.$$
To this aim, observe that by the very definition of $\mathcal G^-_{k^{\star}_n}$, we have
\begin{align*}
\mathbb E\left [ \sum_{A \in \mathcal G^-_{k^{\star}_n}}\hat L_{n}(A)\mu(A)\right] &\leq  \mathbb E \left [ \sum_{A \in \mathcal G^-_{k^{\star}_n}} \left (\hat L_n(A,k^+)+\varphi(A)\right)\mu(A)\right]\\
&=\mathbb E \left [ \sum_{A \in \mathcal G^-_{k^{\star}_n}} \hat L_n(A,k^+)\mu(A)\right]+\mathbb E \left [ \sum_{A \in \mathcal G^-_{k^{\star}_n}} \varphi (A)\mu(A)\right]\\
&  \stackrel{\mbox{\footnotesize def}}{=}\mbox{\sc I}+\mbox{\sc II}.
\end{align*}
For every cell $A$ of $\mathcal G_{k_n^{\star}}^-$, one has 
\begin{equation}
\label{FH2}
\max\left (\frac{n}{2^{dk_n^{\star}}}-1,1\right)\leq N(A)+1 \leq \frac{n}{2^{dk_n^{\star}}}+1.
\end{equation}
Therefore, taking $n$ so large that $n/2^{dk_n^{\star}}> 2$ (this is possible by Proposition \ref{lemmepsi}$(ii)$), we obtain
$$\mbox{\sc II} \leq\left (\frac{n}{2^{dk_n^{\star}}}-1\right)^{-\beta}\mathbb E \left [\sum_{A \in \mathcal G^-_{k^{\star}_n}}\mu(A)\right]\leq \left (\frac{n}{2^{dk_n^{\star}}}-1\right)^{-\beta}.$$
Applying Proposition \ref{lemmepsi}$(ii)$ again, we conclude that $\mbox{\sc II}\to 0$ as $n\to \infty$. 
\medskip

Next, define
$$ k_n=\left \lfloor \alpha \log_2 \left ( \frac{n}{2^{dk^{\star}_n}}-1\right)\right\rfloor \quad \mbox{and} \quad k'_n=\left \lfloor \alpha \log_2 \left ( \frac{n}{2^{dk^{\star}_n}}+1\right)\right\rfloor.$$
Inequality (\ref{FH2}) implies that for every $A\in  \mathcal G^-_{k_n^{\star}}$ and all $n$ large enough,
$$ k_n \leq k^+\leq k'_n.$$
Thus, by Fact \ref{fait1}$(iii)$,
\begin{align*}
\mbox{\sc I}&\leq \mathbb E \left [ \sum_{A \in \mathcal G^-_{k_n^{\star}}}\hat L_n\left (A,k_n\right)\mu(A)\right]+\mathbb E \left [ \sum_{A \in \mathcal G^-_{k_n^{\star}}} \frac{2^{dk'_n}}{N(A)}\mu(A)\right]\\
& = \mathbb E \left [ \sum_{A \in \mathcal G^-_{k_n^{\star}}}\hat L_n\left (A,k_n\right)\mu(A)\right]+ \mbox{O}\left (\left (\frac{2^{dk_n^{\star}}}{n}\right)^{1-d\alpha}\right).
\end{align*}
On the other hand,
\begin{align*}
 & \mathbb E \left [ \sum_{A \in \mathcal G^-_{k_n^{\star}}}\hat L_n\left (A,k_n\right)\mu(A)\right]\\
 & \quad \leq \mathbb E \left [ \sum_{A \in \mathcal G^-_{k_n^{\star}}}L^{\star}(A,k_n)\mu(A)\right] + \mathbb E \left [ \sum_{A \in \mathcal G^-_{k_n^{\star}}}\left |\hat L_n(A,k_n)-L^{\star}(A,k_n)\right|\mu(A)\right]\\
& \quad =\mathbb E \left [ \sum_{A \in \mathcal G^-_{k_n^{\star}}}L^{\star}(A,k_n)\mu(A)\right] + \mbox{O} \left (\sqrt {\left (\frac{2^{dk^{\star}_{n}}}{n}\right)^{1-d\alpha}}\right)\\
& \qquad \mbox{(by Lemma \ref{plus1})}.
\end{align*}
Consequently,
$$I\leq \mathbb E \left [ \sum_{A \in \mathcal G^-_{k_n^{\star}}}L^{\star}(A,k_n)\mu(A)\right]+\mbox{O} \left (\sqrt {\left (\frac{2^{dk^{\star}_{n}}}{n}\right)^{1-d\alpha}}\right),$$
and the rightmost term tends to 0 as $n\to \infty$ by Proposition \ref{lemmepsi}$(ii)$. Thus, the proof will be finalized if we show that
$$ \mathbb E \left [ \sum_{A \in \mathcal G^-_{k_n^{\star}}}L^{\star}(A,k_n)\mu(A)\right] \to L^{\star}\quad \mbox{as }n \to \infty.$$
We have
\begin{align*}
\mathbb E \left [ \sum_{A \in \mathcal G^-_{k_n^{\star}}} L^{\star}(A,k_n)\mu(A)\right]& =\mathbb E \left [ \sum_{A \in \mathcal G^-_{k_n^{\star}}}\sum_{A_j \in \mathcal P_{k_n}(A)}L^{\star}(A_j)\frac{\mu(A_j)}{\mu(A)}\mu(A)\right]\\
& \leq \mathbb E \left [ \sum_{A \in \mathcal P_{k_n}} L^{\star}(A)\mu(A)\right]\\
&=L^{\star}_{k_n},
\end{align*}
where, in the inequality, we use the fact that the cells in the double sum are at level at least $k_n$. 
But, clearly,
$$\frac{{k_n}2^{dk_n}}{n}\leq \frac{\alpha \log_2n}{n^{1-d\alpha}},$$
and consequently, since $d\alpha<1$, 
$$\frac{{k_n}2^{dk_n}}{n}\to 0\quad \mbox{as }n \to \infty.$$
Thus, by Proposition \ref{lemmepsi}$(i)$, the term $L^{\star}_{k_n}$ tends to $L^{\star}$. This concludes the proof.
\section{Some technical results}
Throughout this section, we adopt the general notation of the document. In particular, we let $\alpha$ and $\beta$ be two positive real numbers such that $1-d\alpha-2\beta>0$. The sequence $\left \{k^{\star}_n\right\}_{n \geq 1}$ is defined as in Proposition \ref{lemmepsi} and we set 
\begin{equation}
\label{110812}
k^+= \left \lfloor {\alpha \log_2  (N(A)+1) } \right \rfloor. 
\end{equation}
We will repeatedly use the fact that, by Proposition \ref{lemmepsi}$(ii)$, $2^{dk_n^{\star}}/n \to 0$ as $n\to \infty$. For any $k\geq 0$, $\mathcal T_{k}$ stands for the full $2^{d}$-ary median-type tree with $k$ levels of nodes, whose leaves represent $\mathcal P_k$.
\medskip

Recall that $\bX$ has probability measure $\mu$ on $\mathbb R^d$ and that its marginals are assumed to be nonatomic. The first important result that is needed here is the following one:
\begin{pro}
\label{partition+}
Let $\{k_n\}_{n\geq 1}$ be a sequence of nonnegative integers such that $2^{dk_n}/n\to 0$. Then
$$\mathbb E \left [\sum_{A \in \mathcal P_{k_n}}\left | \frac{N(A)}{n}-\mu(A)\right|\right]=\emph{O}\left (\sqrt{\frac{2^{dk_n}}{n}}\right).$$
\end{pro}
\noindent{\bf Proof of Proposition \ref{partition+}}\quad In the sequel, we let $n$ be large enough to ensure that $n/2^{dk_n}>2$, so that we do not have to worry about empty cells.
\medskip

To prove the lemma, recall the construction of $\mathcal T_{k_n}$. At the root, which represents $\mathbb R^d$, we order the points by the first component. We define the pivot as the $r$-th smallest point, where $r=\lfloor (n+1)/2\rfloor$, and cut 
perpendicularly to the first component at the pivot. 
Let the pivot's first component have value $x^{\star}$. Define
$$A=\left\{\bx \in \mathbb R^d  : \bx=(x_1, \cdots, x_d), x_1<x^{\star}\right\}$$
and
$$B=\left\{\bx \in \mathbb R^d  : \bx=(x_1, \cdots, x_d), x_1>x^{\star}\right\}.$$
The sample points that fall in $A$, conditionally on the pivot, are distributed according to $\mu$ restricted to $A$, and similarly for $B$. Also, importantly,
$$\mu(A)\stackrel{\mathcal L}{=}\mbox{Beta}(r,n-r+1)$$
and
$$\mu(B)\stackrel{\mathcal L}{=}\mbox{Beta}(n-r+1,r),$$
from the theory of order statistics \citep[see, e.g.,][]{DN}.
\medskip

We need to see how large $\mu(A)$, $\mu(B)$, $N(A)$ and $N(B)$ are. To this aim, we distinguish between the cases where $n$ is odd and $n$ is even. 
\begin{enumerate}
\item {\bf $n$ odd}.\quad Now $r=(n+1)/2$, $N(A)=r-1=(n-1)/2$, $N(B)=n-r=(n-1)/2$, and
$$\mu(A)\stackrel{\mathcal L}{=}\mu(B)\stackrel{\mathcal L}{=}\mbox{Beta}\left(\frac{n+1}{2},\frac{n+1}{2}\right).$$
\item {\bf $n$ even}.\quad In this case we have $r=n/2$, $N(A)=(n-2)/2$, $N(B)=n/2$, 
$$\mu(A)\stackrel{\mathcal L}{=}\mbox{Beta}\left(\frac{n}{2},\frac{n+2}{2}\right)$$
and
$$\mu(B)\stackrel{\mathcal L}{=}\mbox{Beta}\left(\frac{n+2}{2},\frac{n}{2}\right).$$
\end{enumerate}
As $N(A)+N(B)=n-1$, the pivot is not sent down to the subtrees. Let us have a canonical way of deciding who goes left and right, e.g., $A$ is left and $B$ is right. 
Next, still at the root, we rotate the coordinate and repeat the median splitting process for the sample points in $A$ and $B$ (both open sets) in direction $2$, then in direction 3, and so forth until direction $d$. We create this way the $2^d$ children of the root and, repeating this scheme for $k_n$ levels of nodes, we construct the $2^d$-ary tree up to distance $k_n$ from the root. It has exactly $2^{dk_n}$ leaves.
\medskip

On any path of length $k_n$ to one of the $2^{dk_n}$ leaves, we have a deterministic sequence of cardinalities
$$n_0=n (\mbox{root}), n_1, n_2, \hdots, n_{k_n}.$$
We have already seen that, for all $i=0, \hdots, k_n$,
$$
\frac{n}{2^{di}}-2\leq n_i \leq \frac{n}{2^{di}}.
$$
Now, consider a fixed path to a fixed leaf, $(n_0, n_1, \hdots, n_{k_n})$. Then, conditionally on the pivots, the set of $\mathbb R^d$ that corresponds to that leaf, i.e., a hyperrectangle  of $\mathbb R^d$, has $\mu$-measure distributed as
\begin{align*}
\mbox{Beta}(n_1+1,n_0-n_1)\times \cdots \times \mbox{Beta}(n_{k_n}+1,n_{{k_n}-1}-n_{k_n})&\stackrel{\mbox{\footnotesize def}}{=}Z_1\times \cdots\times Z_{k_n}\\
& \stackrel{\mbox{\footnotesize def}}{=}Z.
\end{align*}
Observe that 
$$\mathbb E Z=\prod_{i=1}^{k_n}\mathbb E Z_i =\prod_{i=1}^{k_n}\frac{n_i+1}{n_{i-1}+1}=\frac{n_{k_n}+1}{n+1}.$$
Also,
 $$\mathbb E Z^2=\prod_{i=1}^{k_n}\mathbb E Z_i^2 =\prod_{i=1}^{k_n}\frac{(n_i+1)(n_i+2)}{(n_{i-1}+1)(n_{i-1}+2)}=\frac{(n_{k_n}+1)(n_{k_n}+2)}{(n+1)(n+2)}.$$
 The objective is to bound
 \begin{align*}
 \mathbb E \left | \frac{n_{k_n}}{n}-Z\right| &\leq  \sqrt{\mathbb E \left |Z- \frac{n_{k_n}}{n}\right|^2}\\
 & =\sqrt{\mathbb E \left |Z- \mathbb E Z\right|^2+ \left |\frac{n_{k_n}}{n}- \mathbb E Z\right|^2}\\
 & =\sqrt{\mathbb V Z+ \left |\frac{n_{k_n}}{n}- \mathbb E Z\right|^2}\\
 & =\sqrt{\mathbb V Z+ \left |\frac{n_{k_n}}{n}- \frac{n_{k_n}+1}{n+1}\right|^2},
 \end{align*}
 where the symbol $\mathbb V$ stands for the variance. Note
 $$\left | \frac{n_{k_n}}{n}-\frac{n_{k_n}+1}{n+1}\right|=\left |\frac{n_{k_n}-n}{n(n+1)}\right|\leq \frac{1}{n+1}.$$
 Also,
 \begin{align*}
 \mathbb V Z &=\left (\frac{n_{k_n}+1}{n+1}\right)\left (\frac{n_{k_n}+2}{n+2}-\frac{n_{k_n}+1}{n+1}\right)\\
 &=\left (\frac{n_{k_n}+1}{n+1}\right)\times \frac{n-n_{k_n}}{(n+2)(n+1)}\\
 & \leq \frac{n_{k_n}+1}{(n+1)(n+2)}.
 \end{align*}
Thus, 
\begin{align*}
\mathbb E \left | \frac{n_{k_n}}{n}-Z\right| &\leq \sqrt{\frac{n_{k_n}+1}{(n+1)(n+2)}+\frac{1}{(n+1)^2}}\\
& \leq \frac{1}{n+1}\sqrt{n_{k_n}+2}.
\end{align*}
Sum over all $2^{d{k_n}}$ sets in the partition $\mathcal P_{k_n}$, and call the 
set cardinalities $n_{k_n}(1), \hdots, n_{k_n}(2^{dk_n})$. Then, denoting by $Z_i$ the ``$Z$'' for the $i$-th set in the partition, we obtain
\begin{align*}
\mathbb E \left [\sum_{i=1}^{2^{d{k_n}}} \left | \frac{n_{k_n}(i)}{n}-Z_i\right|\right] & \leq \frac{1}{n+1} \sum_{i=1}^{2^{dk_n}} \sqrt{n_{k_n}(i)+2}\\
& \leq \frac{1}{n+1}\sqrt{\sum_{i=1}^{2^{dk_n}}1}\sqrt{\sum_{i=1}^{2^{dk_n}}(n_{k_n}(i)+2)}\\
& \quad (\mbox{by the Cauchy-Schwarz inequality}).
\end{align*}
Therefore,
\begin{align*}
\mathbb E \left [\sum_{i=1}^{2^{dk_n}} \left | \frac{n_{k_n}(i)}{n}-Z_i\right|\right] & \leq \frac{\sqrt{2^{dk_n}}}{n+1}\times \sqrt{n+2^{{dk_n}+1}}\\
&\leq \frac{\sqrt{2^{dk_n}}}{n+1} \left ( \sqrt n+\sqrt{2^{{dk_n}+1}}\right)\\
& \leq \sqrt{\frac{2^{dk_n}}{n}} +\frac{2^{{dk_n}+1}}{n}.
\end{align*}
Since $2^{dk_n}/n \to 0$ as $n\to \infty$, this last term
is $\mbox{O}(\sqrt{2^{dk_n}/n})$.
\hfill $\blacksquare$
\begin{cor}
\label{partition+SUB}
Let $\{k_n\}_{n \geq 1}$ be a sequence of nonnegative integers such that $2^{dk_n}/n \to 0$, and let $\mathcal P^-_{k_n}$ be the partition of $\mathbb R^d$ corresponding to the leaves of any subtree of $\mathcal T_{k_n}$ rooted at $\mathbb R^d$. Then
$$\mathbb E \left [\sum_{A \in \mathcal P^-_{k_n}}\left | \frac{N(A)}{n}-\mu(A)\right|\right]=\emph{O}\left (\sqrt{\frac{2^{dk_n}}{n}}\right).$$
\end{cor}
\noindent{\bf Proof of Corollary \ref{partition+SUB}}\quad The proof is similar to the proof of Proposition \ref{partition+}---just note that $\mathcal P_{k_n}^-$ has at most $2^{dk_n}$ cells.
\hfill $\blacksquare$
\begin{pro}
\label{castle1}
Let $\{k_n\}_{n \geq 1}$ be a sequence of nonnegative integers such that $2^{dk_n}/n\to 0$. Then
\begin{align*}
&\mathbb E \left |\frac{1}{N\left(A_{k_n}(\bX)\right)}\sum_{i=1}^n \mathbf 1_{[\bX_i \in A_{k_n}(\bX),Y_i=1]}-\frac{1}{\mu\left(A_{k_n}(\bX)\right)}\int_{A_{k_n}(\bX)} \eta(\bz)\mu(\emph{d} \bz)\right|\\
& \quad =\emph{O}\left (\sqrt{\frac{2^{dk_n}}{n}}\right)
\end{align*}
and, similarly,
\begin{align*}
&\mathbb E \left |\frac{1}{N\left(A_{k_n}(\bX)\right)}\sum_{i=1}^n \mathbf 1_{[\bX_i \in A_{k_n}(\bX),Y_i=0]}-\frac{1}{\mu\left(A_{k_n}(\bX)\right)}\int_{A_{k_n}(\bX)} \left (1-\eta(\bz)\right)\mu(\emph{d} \bz)\right|\\
& \quad =\emph{O}\left (\sqrt{\frac{2^{dk_n}}{n}}\right).
\end{align*}
\end{pro}
\noindent{\bf Proof of Proposition \ref{castle1}}\quad  We only prove the first statement. Since $n/2^{dk_n}\to \infty$ as $n \to \infty$, we can always choose $n$ large enough so that no cell of $\mathcal P_{k_n}$ is empty. A quick check of $\mathcal T_{k_n}$ reveals that given the pivots (see Proposition \ref{partition+}), the points inside each cell are distributed in an i.i.d.~manner according  to the restriction of $\mu$ to the cell. Moreover, conditionally on $\bX$ and the pivots, $N(A_{k_n}(\bX))$ has a deterministic, fixed value. Thus, setting
$$\bar \eta_{n}(\bx)=\frac{1}{\mu\left(A_{k_n}(\bx)\right)}\int_{A_{k_n}(\bx)}\eta(\bz)\mu(\mbox{d}\bz),$$
we obtain, conditionally on $\bX$ and the pivots, 
\begin{align*}
&\mathbb E\left |\frac{1}{N\left(A_{k_n}(\bX)\right)}\sum_{i=1}^n \mathbf 1_{[\bX_i \in A_{k_n}(\bX),Y_i=1]}-\frac{1}{\mu\left(A_{k_n}(\bX)\right)}\int_{A_{k_n}(\bX)} \eta(\bz)\mu(\mbox{d} \bz)\right|\\
&\quad \leq \sqrt{\frac{\bar \eta_{n}(\bX)\left(1-\bar \eta_n(\bX)\right)}{N\left(A_{k_n}(\bX)\right) }}\\
& \quad \leq \frac{1}{2}\sqrt{\frac{1}{N\left(A_{k_n}(\bX)\right) }}\\
& \quad \leq \frac{1}{2}\sqrt{\frac{1}{\frac{n}{2^{dk_n}}-2}}.
\end{align*}
The result follows from the condition $2^{dk_n}/n \to 0$.
\hfill $\blacksquare$
\begin{cor}
\label{castle2}
Let $\{k_n\}_{n \geq 1}$ be a sequence of nonnegative integers such that $2^{dk_n}/n\to 0$ , and let $\mathcal P^-_{k_n}$ be the partition of $\mathbb R^d$ corresponding to the leaves of any subtree of $\mathcal T_{k_n}$ rooted at $\mathbb R^d$. For each $\bx \in \mathbb R^d$, denote by $A^-_{k_n}(\bx)$ the cell of $\mathcal P^-_{k_n}$ containing $\bx$. Then
\begin{align*}
&\mathbb E \left |\frac{1}{N\left(A^-_{k_n}(\bX)\right)}\sum_{i=1}^n \mathbf 1_{[\bX_i \in A^-_{k_n}(\bX),Y_i=1]}-\frac{1}{\mu\left(A^-_{k_n}(\bX)\right)}\int_{A^-_{k_n}(\bX)} \eta(\bz)\mu(\emph{d} \bz)\right|\\
& \quad =\emph{O}\left (\sqrt{\frac{2^{dk_n}}{n}}\right)
\end{align*}
and, similarly,
\begin{align*}
&\mathbb E \left |\frac{1}{N\left(A^-_{k_n}(\bX)\right)}\sum_{i=1}^n \mathbf 1_{[\bX_i \in A^-_{k_n}(\bX),Y_i=0]}-\frac{1}{\mu\left(A^-_{k_n}(\bX)\right)}\int_{A^-_{k_n}(\bX)} \left (1-\eta(\bz)\right)\mu(\emph{d} \bz)\right|\\
& \quad =\emph{O}\left (\sqrt{\frac{2^{dk_n}}{n}}\right).
\end{align*}
\end{cor}
\noindent{\bf Proof of Corollary \ref{castle2}}\quad The proof is similar to that of Proposition \ref{castle1}---just note that
$$N\left(A^-_{k_n}(\bX)\right)\geq \frac{n}{2^{dk_n}}$$
for all $n$ large enough.
\hfill $\blacksquare$
\begin{lem}
\label{lemmepartition}
Let $\{k_n\}_{n \geq 1}$ be a sequence of nonnegative integers such that $2^{dk_n}/n$ $\to 0$. Then
$$\mathbb E \left |\hat L_{n}\left (A_{k_n}(\bX)\right)-L^{\star}\left(A_{k_n}(\bX)\right)\right| =\emph{O}\left (\sqrt{\frac{2^{dk_n}}{n}}\right).$$
\end{lem}
\noindent{\bf Proof of Lemma \ref{lemmepartition}}\quad Using the definition of $\hat L_{n}(A_{k_n}(\bX))$ and $L^{\star}(A_{k_n}(\bX))$, we may write
\begin{align*}
&\mathbb E\left |\hat L_{n}\left (A_{k_n}(\bX)\right)-L^{\star}\left(A_{k_n}(\bX)\right)\right|\\
&  \leq\mathbb E\left |\frac{1}{N\left(A_{k_n}(\bX)\right)}\sum_{i=1}^n \mathbf 1_{[\bX_i \in A_{k_n}(\bX),Y_i=1]}-\frac{1}{\mu\left(A_{k_n}(\bX)\right)}\int_{A_{k_n}(\bX)} \eta(\bz)\mu(\mbox{d} \bz)\right|\\
& \quad + \mathbb E\Bigg |\frac{1}{N\left(A_{k_n}(\bX)\right)}\sum_{i=1}^n \mathbf 1_{[\bX_i \in A_{k_n}(\bX),Y_i=0]}\\
& \quad \qquad \qquad -\frac{1}{\mu\left(A_{k_n}(\bX)\right)}\int_{A_{k_n}(\bX)} \left (1-\eta(\bz)\right)\mu(\mbox{d} \bz)\Bigg|.
\end{align*}
Each term of the sum goes to 0 by Proposition \ref{castle1}.
\hfill $\blacksquare$
\begin{lem}
\label{lemmepartitionSUB}
Let $\{k_n\}_{n \geq 1}$ be a sequence of nonnegative integers such that $2^{dk_n}/n$ $\to 0$, and let $\mathcal P^-_{k_n}$ be the partition of $\mathbb R^d$ corresponding to the leaves of any subtree of $\mathcal T_{k_n}$ rooted at $\mathbb R^d$. For each $\bx \in \mathbb R^d$, denote by $A^-_{k_n}(\bx)$ the cell of $\mathcal P^-_{k_n}$ containing $\bx$.  Then
$$\mathbb E \left | \hat L_n \left (A^-_{k_n}(\bX)\right) -L^{\star} \left (A^-_{k_n}(\bX)\right)\right|=\emph{O} \left (\sqrt{\frac{2^{dk_n}}{n}}\right).$$
\end{lem}
\noindent{\bf Proof of Lemma \ref{lemmepartitionSUB}}\quad The proof is similar to that of Lemma \ref{lemmepartition}. It uses Corollary \ref{castle2} instead of Proposition \ref{castle1}.
\hfill $\blacksquare$
\begin{lem}
\label{LccBIS}
Let
$$k_n= \left \lfloor \alpha \log_2 \left ( \frac{n}{2^{dk^{\star}_n}}+1\right)\right\rfloor.$$
Then
$$\mathbb E \left |\hat L_n(A_{k_n^{\star}}(\bX),k_n)-L^{\star}(A_{k_n^{\star}}(\bX),k_n) \right|=\emph {O} \left (\sqrt{\left(\frac{2^{dk_n^{\star}}}{n}\right)^{1-d\alpha}}\right).$$
\end{lem}
\noindent{\bf Proof of Lemma \ref{LccBIS}}\quad We have
\begin{align*}
&\mathbb E \left |\hat L_n(A_{k_n^{\star}}(\bX),k_n)-L^{\star}(A_{k_n^{\star}}(\bX),k_n) \right| \\
& \quad = \mathbb E \left [ \sum_{A \in \mathcal P_{k_n^{\star}}} \sum_{A_j \in \mathcal P_{k_n}(A)} \left |\hat L_n(A_j)\frac{N(A_j)}{N(A)}- L^{\star}(A_j)\frac{\mu(A_j)}{\mu(A)}\right|\mu(A)\right]\\
& \quad \leq \mathbb E \left [ \sum_{A \in \mathcal P_{k_n^{\star}}} \sum_{A_j \in \mathcal P_{k_n}(A)} \left |\hat L_n(A_j)\frac{N(A_j)}{N(A)}- \hat L_n(A_j)\frac{\mu(A_j)}{\mu(A)}\right|\mu(A)\right]\\
& \qquad +\mathbb E \left [ \sum_{A \in \mathcal P_{k_n^{\star}}} \sum_{A_j \in \mathcal P_{k_n}(A)} \left |\hat L_n(A_j)- L^{\star}(A_j)\right|\mu(A_j)\right]\\
& \quad  \stackrel{\mbox{\footnotesize def}}{=}\mbox{\sc I}+\mbox{\sc II}.
\end{align*}
Clearly,
$$\mbox{\sc II} =\mathbb E \left | \hat L_n \left (A_{k_n^{\star}+k_n}(\bX)\right) -L^{\star} \left (A_{k_n^{\star}+k_n}(\bX)\right)\right|$$
whence, according to Lemma \ref{lemmepartition},
$$\mbox{\sc II} =\mbox {O} \left (\sqrt{\left(\frac{2^{dk_n^{\star}}}{n}\right)^{1-d\alpha}}\right).$$
On the other hand, since $\hat L_n(A_j)\leq 1$,
\begin{align*}
\mbox{\sc I} & \leq \mathbb E \left [  \sum_{A \in \mathcal P_{k_n^{\star}}} \sum_{A_j \in \mathcal P_{k_n}(A)} \left |\frac{N(A_j)}{N(A)}- \frac{\mu(A_j)}{\mu(A)}\right|\mu(A)\right]\\
& \leq\mathbb E \left [  \sum_{A \in \mathcal P_{k_n^{\star}}} \sum_{A_j \in \mathcal P_{k_n}(A)} \left |\frac{N(A_j)}{N(A)}\mu(A)- \frac{N(A_j)}{n}\right|\right]\\
&  \quad + \mathbb E \left [  \sum_{A \in \mathcal P_{k_n^{\star}}} \sum_{A_j \in \mathcal P_{k_n}(A)} \left |\frac{N(A_j)}{n}-\mu(A_j)\right|\right].
\end{align*}
The inequality
$$\sum_{A_j \in \mathcal P_{k_n}(A)} N(A_j)\leq N(A)$$
leads to 
\begin{align*}
\mbox{\sc I} & \leq \mathbb E \left [  \sum_{A \in \mathcal P_{k_n^{\star}}}  \left |\mu(A)-\frac{N(A)}{n}\right|\right]+ \mathbb E \left [  \sum_{A \in \mathcal P_{k_n^{\star}}} \sum_{A_j \in \mathcal P_{k_n}(A)} \left |\frac{N(A_j)}{n}-\mu(A_j)\right|\right]\\
&= \mathbb E \left [  \sum_{A \in \mathcal P_{k_n^{\star}}}  \left |\mu(A)-\frac{N(A)}{n}\right|\right]+ \mathbb E \left [  \sum_{A \in \mathcal P_{k_n^{\star}+k_n}} \left |\frac{N(A)}{n}-\mu(A)\right|\right].
\end{align*}
Thus, by Proposition \ref{partition+},
$$
\mbox{\sc I} =\mbox {O} \left (\sqrt{\left(\frac{2^{dk_n^{\star}}}{n}\right)^{1-d\alpha}}\right).
$$
Collecting bounds, we obtain
$$\mbox{\sc I}+\mbox{\sc II}=\mbox {O} \left (\sqrt{\left(\frac{2^{dk_n^{\star}}}{n}\right)^{1-d\alpha}}\right).$$
\hfill $\blacksquare$
\begin{lem}
\label{plus1}
Let $\mathcal G_{k_n^{\star}}^-$ be the collection of cells of $\mathcal G_n$ at level at most $k^{\star}_n$, and let
$$k_n=\left \lfloor \alpha \log_2 \left ( \max\left(\frac{n}{2^{dk^{\star}_n}}-1\right),1\right)\right\rfloor.$$
Then
$$
\mathbb E \left [ \sum_{A \in \mathcal G^-_{k_n^{\star}}}\left |\hat L_n(A,k_n)-L^{\star}(A,k_n)\right|\mu(A)\right]=\emph{O} \left (\sqrt {\left (\frac{2^{dk^{\star}_{n}}}{n}\right)^{1-d\alpha}}\right).$$
\end{lem}
\noindent{\bf Proof of Lemma \ref{plus1}}\quad
Denote by $\bar {\mathcal G}^-_{k_n^{\star}}$ the cells of $\mathcal P_{k_n^{\star}}$ such that the path from the root to the cell does not cross ${\mathcal G}^-_{k_n^{\star}}$. By construction, the subset collection
$$\mathcal P^-_{k_n^{\star}}=\mathcal G^-_{k_n^{\star}}Ê\cup \bar {\mathcal G}^-_{k_n^{\star}}$$
is a partition of $\mathbb R^d$ represented by a subtree of $\mathcal T_{k_n^{\star}}$ rooted at $\mathbb R^d$. Moreover, clearly,
\begin{align*}
& \mathbb E \left [ \sum_{A \in \mathcal G^-_{k_n^{\star}}}\left |\hat L_n(A,k_n)-L^{\star}(A,k_n)\right|\mu(A)\right]\\
&\quad \leq  \mathbb E \left [ \sum_{A \in \mathcal P^-_{k_n^{\star}}}\left |\hat L_n(A,k_n)-L^{\star}(A,k_n)\right|\mu(A)\right].
\end{align*}
Thus, denoting by $A_{k_n^{\star}}^-(\bx)$ the cell of $\mathcal P^-_{k^{\star}_n}$ containing $\bx$, we are led to
\begin{align*}
& \mathbb E \left [ \sum_{A \in \mathcal G^-_{k_n^{\star}}}\left |\hat L_n(A,k_n)-L^{\star}(A,k_n)\right|\mu(A)\right]\\
&\quad \leq  \mathbb E \left |\hat L_n(A^-_{k_n^{\star}}(\bX),k_n)-L^{\star}(A^-_{k_n^{\star}}(\bX),k_n) \right|.
\end{align*}
The end of the proof is similar to the proof of Lemma \ref{LccBIS}. Replace $\mathcal P_{k_n^{\star}}$ by $\mathcal P^-_{k_n^{\star}}$ and invoke Corollary \ref{partition+SUB} (instead of Proposition \ref{partition+}) and Lemma \ref{lemmepartitionSUB} (instead of Lemma \ref{lemmepartition}).
\hfill $\blacksquare$
\begin{pro}
\label{Lcc}
Let $k^+$ be defined as in (\ref{110812}). Then
$$\mathbb E \left |\hat L_{n}\left (A_{k_n^{\star}}(\bX)\right)-\hat L_{n} (A_{k_n^{\star}}(\bX),k^+) \right| \leq \psi(n,k_n^{\star})+\emph{O} \left ( \sqrt{\left(\frac{2^{dk_n^{\star}}}{n}\right)^{1-d\alpha}}\right),$$
where
$$\psi(n,k)=L_{k}^{\star}-L^{\star}.$$
\end{pro}
\noindent{\bf Proof of Proposition \ref{Lcc}}\quad 
For every cell $A$ of $\mathcal P_{k_n^{\star}}^-$, one has 
\begin{equation}
\label{FH3}
\max\left (\frac{n}{2^{dk_n^{\star}}}-1,1\right)\leq N(A)+1 \leq \frac{n}{2^{dk_n^{\star}}}+1.
\end{equation}
Define
$$ k'_n=\left \lfloor \alpha \log_2 \left ( \frac{n}{2^{dk^{\star}_n}}-1\right)\right\rfloor \quad \mbox{and} \quad k_n=\left \lfloor \alpha \log_2 \left ( \frac{n}{2^{dk^{\star}_n}}+1\right)\right\rfloor,$$
and note that, by inequalities (\ref{FH3}), for all $n$ large enough,
$$ k'_n \leq k^+\leq k_n.$$
Thus, by the triangle inequality and Fact \ref{fait1}$(ii)$, we may write
\begin{align*}
&\mathbb E \left |\hat L_{n}\left (A_{k_n^{\star}}(\bX)\right)-\hat L_{n} (A_{k_n^{\star}}(\bX),k^+) \right| \\
& \quad \leq \mathbb E \left [\hat L_{n}\left (A_{k_n^{\star}}(\bX)\right)-\hat L_{n} (A_{k_n^{\star}}(\bX),k^+) + \frac{2^{dk^+}}{N\left(A(\bX)\right)}\mathbf 1_{[N(A(\bX))>0]}\right]\\
&\qquad + \mathbb E \left [Ê\frac{2^{dk^+}}{N\left(A(\bX)\right)}\mathbf 1_{[N(A(\bX))>0]}\right]\\
& \quad = \mathbb E \left [\hat L_{n}\left (A_{k_n^{\star}}(\bX)\right)-\hat L_{n} (A_{k_n^{\star}}(\bX),k^+) \right] +\mbox{O} \left ( {\left(\frac{2^{dk_n^{\star}}}{n}\right)^{1-d\alpha}}\right).
\end{align*}
Consequently, by Fact \ref{fait1}$(iii)$,
\begin{align}
&\mathbb E \left |\hat L_{n}\left (A_{k_n^{\star}}(\bX)\right)-\hat L_{n} (A_{k_n^{\star}}(\bX),k^+) \right| \nonumber\\
& \quad \leq \mathbb E \left [\hat L_{n}\left (A_{k_n^{\star}}(\bX)\right)-\hat L_{n} (A_{k_n^{\star}}(\bX),k_n)\right]+\mbox{O} \left ( {\left(\frac{2^{dk_n^{\star}}}{n}\right)^{1-d\alpha}}\right). \label{allez}
\end{align}
With respect to the first term on the right-hand side, we have
\begin{align*}
&\mathbb E \left [\hat L_{n}\left (A_{k_n^{\star}}(\bX)\right)-\hat L_{n} (A_{k_n^{\star}}(\bX),k_n)\right]\\
&\quad  \leq \mathbb E \left |\hat L_{n}\left (A_{k_n^{\star}}(\bX)\right)-L^{\star}\left (A_{k_n^{\star}}(\bX)\right)\right| \\
& \qquad +L^{\star}_{k_n^{\star}}-L^{\star}\\
& \qquad + L^{\star}-\mathbb E \left [L^{\star}(A_{k_n^{\star}}(\bX),k_n)\right]\\
& \qquad +\mathbb E \left |L^{\star}(A_{k_n^{\star}}(\bX),k_n)-\hat L_n(A_{k_n^{\star}}(\bX),k_n)\right|.
\end{align*}
According to Lemma \ref{lemmepartition}, the first of the four terms above is $\mbox{O}(\sqrt{2^{dk_n^{\star}}/n})$, whereas the third one is nonpositive by Fact \ref{fait1}$(iv)$. Consequently, 
\begin{align*}
&\mathbb E \left [\hat L_{n}\left (A_{k_n^{\star}}(\bX)\right)-\hat L_{n} \left (A_{k_n^{\star}}(\bX),k_n\right) \right] \leq \psi(n,k_n^{\star})+\mbox{O} \left ( \sqrt{\frac{2^{dk_n^{\star}}}{n}}\right)\\
& \quad +\mathbb E \left |L^{\star}(A_{k_n^{\star}}(\bX),k_n )-\hat L_n(A_{k_n^{\star}}(\bX),k_n)\right|.
\end{align*}
Evoking finally Lemma \ref{LccBIS}, we see that
$$\mathbb E \left [\hat L_{n}\left (A_{k_n^{\star}}(\bX)\right)-\hat L_{n} (A_{k_n^{\star}}(\bX),k_n) \right] \leq \psi(n,k_n^{\star})+\mbox{O} \left ( \sqrt{\left(\frac{2^{dk_n^{\star}}}{n}\right)^{1-d\alpha}}\right).$$
Combining this result with (\ref{allez}) leads to the desired statement.
\hfill $\blacksquare$
\begin{lem}
\label{greedymoins}
Let $k^+$ be defined as in (\ref{110812}). Then
\begin{align*}
&\mathbb P \left \{ \left |\hat L_{n}\left(A_{k_n^{\star}}(\bX)\right) -\hat L_{n}(A_{k^{\star}_n}(\bX),k^+)\right|>\left(\frac{1}{N\left(A_{k_n^{\star}}(\bX)\right)+1}\right)^{\beta}\right\}\\
&\quad =\emph{O} \left (\sqrt{ \left (\frac{2^{dk_n^{\star}}}{n}\right)^{1-d\alpha-2\beta}}\right).
\end{align*}
\end{lem}
\noindent{\bf Proof of Lemma \ref{greedymoins}}\quad Set
$$\varphi(A)=\left(\frac{1}{N(A)+1}\right)^{\beta}.$$
Since $N(A_{k_n^{\star}}(\bx))\leq n/2^{dk_n^{\star}}$, 
one has
\begin{align*}
&\mathbb P \left \{\left |\hat L_{n}\left(A_{k_n^{\star}}(\bX)\right) -\hat L_{n}(A_{k_n^{\star}}(\bX),k^+)\right|>\varphi \left (A_{k_n^{\star}}(\bX)\right)\right\}\\
& \quad \leq \mathbb P \left \{\left |\hat L_{n}\left(A_{k_n^{\star}}(\bX)\right) -\hat L_{n}(A_{k_n^{\star}}(\bX),k^+)\right|>\left (\frac{1}{n/2^{dk_n^{\star}}+1}\right)^{\beta}\right\}.
\end{align*}
Therefore, by Markov's inequality, 
\begin{align*}
&\mathbb P \left \{\left|\hat L_{n}\left(A_{k_n^{\star}}(\bX)\right) -\hat L_{n}(A_{k_n^{\star}}(\bX),k^+)\right|>\varphi \left (A_{k_n^{\star}}(\bX)\right)\right\}\\
& \quad \leq  (n/2^{dk_n^{\star}}+1)^{\beta}\times\mathbb E \left |\hat L_{n}\left(A_{k_n^{\star}}(\bX)\right) - \hat L_{n}(A_{k_n^{\star}}(\bX),k^+)\right|.
\end{align*}
Thus, by Proposition \ref{Lcc},
\begin{align*}
&\mathbb P \left \{\left |\hat L_{n}\left(A_{k_n^{\star}}(\bX)\right) -\hat L_{n}(A_{k_n^{\star}}(\bX),k^+)\right|>\varphi \left (A_{k_n^{\star}}(\bX)\right)\right\}\\
& \quad \leq (n/2^{dk_n^{\star}}+1)^{\beta}\times \left [\psi(n,k_n^{\star})+\mbox{O} \left ( \sqrt{\left(\frac{2^{dk_n^{\star}}}{n}\right)^{1-d\alpha}}\right) \right].
\end{align*}
But, by definition of $k^{\star}_n$,
$$\psi(n,k_n^{\star})<\sqrt{\left(\frac{2^{dk^{\star}_n}}{n}\right)^{1-d\alpha}}.$$
It follows, since $n/2^{dk_n^{\star}} \to \infty$, that
$$\mathbb P \left \{\left |\hat L_{n}\left(A_{k_n^{\star}}(\bX)\right) -\hat L_{n}(A_{k^{\star}_n}(\bX),k^+)\right|>\varphi(A)\right\}=\mbox{O} \left (\sqrt{ \left (\frac{2^{dk_n^{\star}}}{n}\right)^{1-d\alpha-2\beta}}\right).
$$
\hfill $\blacksquare$
\begin{lem}
\label{greedyplus} Let $\mathcal G_{k_n^{\star}}^-$ be the collection of cells of $\mathcal G_n$ at level at most $k^{\star}_n$. For $\bx \in \mathbb R^d$, denote by $A_{k_n^{\star}}^-(\bx)$ the cell of $\mathcal G^-_{k^{\star}_n}$ containing $\bx$, and set $N(A_{k_n^{\star}}^-(\bx))=\sum_{i=1}^n \mathbf 1_{[\bX_i \in A_{k_n^{\star}}^-(\bx)]}$. Define
$$\hat \eta_n(\bx)=\frac{1}{N\big(A_{k_n^{\star}}^-(\bx)\big)}\sum_{i=1}^n \mathbf 1_{[\bX_i \in A_{k_n^{\star}}^-(\bx),Y_i=1]}.$$
Then
$$\mathbb E \left [\sum_{A \in \mathcal G^-_{k^{\star}_n}} \left | \int_A \hat \eta_n(\bz)\mu(\emph{d}\bz)-\int_A \eta(\bz)\mu(\emph{d}\bz)\right|\right] =\emph{O}\left ( \sqrt{\frac{2^{dk^{\star}_n}}{n}}\right)$$
and, similarly,
$$\mathbb E \left [\sum_{A \in \mathcal G^-_{k^{\star}_n}} \left | \int_A \left (1-\hat \eta_n(\bz)\right)\mu(\emph{d}\bz)-\int_A \left (1-\eta(\bz)\right)\mu(\emph{d}\bz)\right|\right] =\emph{O} \left (\sqrt{\frac{2^{dk^{\star}_n}}{n}}\right).$$
\end{lem}
\noindent{\bf Proof of Lemma \ref{greedyplus}}\quad We only have to prove the first statement. To this aim, observe that
\begin{align*}
&\mathbb E \left [\sum_{A \in \mathcal G^-_{k^{\star}_n}} \left | \int_A \hat \eta_n(\bz)\mu(\mbox{d}\bz)-\int_A \eta(\bz)\mu(\mbox{d}\bz)\right|\right]\\
& \quad =\mathbb E \left [ \sum_{A \in \mathcal G^-_{k^{\star}_n}} \left |\frac{1}{N(A)}\sum_{i=1}^n \mathbf 1_{[\bX_i\in A, Y_i=1]}-\frac{1}{\mu(A)}\int_A \eta(\bz)\mu(\mbox{d}\bz)\right|\mu(A)\right].
\end{align*}
Denote by $\bar {\mathcal G}^-_{k_n^{\star}}$ the cells of $\mathcal P_{k_n^{\star}}$ such that the path from the root to the cell does not cross ${\mathcal G}^-_{k_n^{\star}}$. By construction, the subset collection
$$\mathcal P^-_{k_n^{\star}}=\mathcal G^-_{k_n^{\star}}Ê\cup \bar {\mathcal G}^-_{k_n^{\star}}$$
is a partition of $\mathbb R^d$ represented by a subtree of $\mathcal T_{k_n^{\star}}$ rooted at $\mathbb R^d$. Now, 
\begin{align*}
&\mathbb E \left [\sum_{A \in \mathcal G^-_{k^{\star}_n}} \left | \int_A \hat \eta_n(\bz)\mu(\mbox{d}\bz)-\int_A \eta(\bz)\mu(\mbox{d}\bz)\right|\right]\\
& \quad \leq\mathbb E \left [ \sum_{A \in \mathcal P^-_{k^{\star}_n}} \left |\frac{1}{N(A)}\sum_{i=1}^n \mathbf 1_{[\bX_i\in A, Y_i=1]}-\frac{1}{\mu(A)}\int_A \eta(\bz)\mu(\mbox{d}\bz)\right|\mu(A)\right].
\end{align*}
But, since $\mathcal P^-_{k_n^{\star}}$ is a partition of $\mathbb R^d$, one has
\begin{align*}
&\mathbb E \left [ \sum_{A \in \mathcal P^-_{k^{\star}_n}} \left |\frac{1}{N(A)}\sum_{i=1}^n \mathbf 1_{[\bX_i\in A, Y_i=1]}-\frac{1}{\mu(A)}\int_A \eta(\bz)\mu(\mbox{d}\bz)\right|\mu(A)\right]\\
&= \mathbb E \left [\left |\frac{1}{N\big(A^-_{k^{\star}_n}(\bX)\big)}\sum_{i=1}^n \mathbf 1_{[\bX_i \in A^-_{k_n^{\star}}(\bX),Y_i=1]}-\frac{1}{\mu\big(A^-_{k^{\star}_n}(\bX)\big)}\int_{A^-_{k^{\star}_n}(\bX)} \eta(\bz)\mu(\mbox{d} \bz)\right|\right].
\end{align*}
This term goes to 0 by Corollary \ref{castle2}.
\hfill $\blacksquare$
\paragraph{Acknowledgments} We thank the Editor and an anonymous referee for valuable comments and insightful suggestions. 
\bibliography{biblio-greedy}
\end{document}